\pdfoutput=1
\documentclass[11pt]{article}

\usepackage{acl} % [review]

\usepackage{times}
\usepackage{latexsym}

\usepackage[T1]{fontenc}
\usepackage[utf8]{inputenc}

\usepackage{microtype}

\usepackage{inconsolata}

\usepackage{multirow}
\usepackage{graphicx}
\usepackage{booktabs}
\usepackage{tabularx}
\usepackage{makecell}
\usepackage{soul}
\usepackage{caption}
\usepackage{subcaption}

\usepackage{tikz}      
\usepackage{ifthen}
\usetikzlibrary{matrix,calc}
\usepackage{xstring}

\usepackage{amsmath}

\title{Sorted LLaMA: Unlocking the Potential of Intermediate Layers of Large Language Models for Dynamic Inference} %  \st{Using Sorted Fine-Tuning (SoFT)}

\author{
    Parsa Kavehzadeh\textsuperscript{\rm 2},
    Mojtaba Valipour\textsuperscript{\rm 1},
    Marzieh Tahaei\textsuperscript{\rm 2}, \\
    {\bf Ali Ghodsi}\textsuperscript{\rm 1},
    {\bf Boxing Chen}\textsuperscript{\rm 2}, and
    {\bf Mehdi Rezagholizadeh}\textsuperscript{\rm 2} \\
    \textsuperscript{\rm 1}University of Waterloo\\
    \textsuperscript{\rm 2}Huawei Noah's Ark Lab\\
    \{mojtaba.valipour, ali.ghodsi\}@uwaterloo.ca, \\
    \{parsa.kavehzadeh, mehdi.rezagholizadeh, marzieh.tahaei, boxing.chen\}@huawei.com
}

\begin{document}

\maketitle

% \textbf{** Color code used for addressing the comments of different reviewers:}
 % \rone{Reviewer 1}, 
 % \rtwo{Reviewer 2}, 
 % \rthree{Reviewer 3},
 % \rgen{General}.

\begin{abstract}
Large language models (LLMs) have revolutionized natural language processing (NLP) by excelling at understanding and generating human-like text. However, their widespread deployment can be prohibitively expensive. SortedNet is a recent training technique for enabling dynamic inference by leveraging the modularity in networks and sorting sub-models based on computation/accuracy in a nested manner. %It leverages network modularity to create sub-models with varying computational loads,  sorting them based on computation/accuracy characteristics in a nested manner. 
We extend SortedNet to generative NLP tasks, making large language models dynamic without any Pre-Training and by only replacing Standard Fine-Tuning (SFT) with Sorted Fine-Tuning (SoFT). Our approach boosts model efficiency, eliminating the need for multiple models for various scenarios during inference. We show that this approach can unlock the power of intermediate layers of transformers in generating the target output. Our sub-models remain integral components of the original model, minimizing storage requirements and transition costs between different computational/latency budgets. The efficacy of our proposed method was demonstrated by applying it to tune LLaMA 2 13B on the Stanford Alpaca dataset for instruction following and TriviaQA for closed-book question answering. Our results show the superior performance of sub-models in comparison to Standard Fine-Tuning and SFT+ICT (Early-Exit), all achieved with efficient tuning and without additional memory usage during inference.
\end{abstract}

\section{Introduction}
%Large Language Models (LLMs) \cite{gpt3} are powerful models that can sovle many down-stream tasks at once. This will make them effective for solving many different problems such as translations \cite{translation}, customer support \cite{}, and even code development \cite{}. One even can fine-tune these large language models to make them strong instruction-followers \cite{Ouyang2022TrainingLM} and eventually teaching them to become helpful assistants. 

%There is an ongoing competition between many industry competitors that trying to train the largest language model possible as larger language models shown better reasoning capabilities and generalization across multiple tasks. There are several papers in the literature that claim emergence capability of these large language models. 

%We also know that these LLMs are modular by design. In other words, there are several transformers blocks \cite{Vaswani2017AttentionIA} that forms a single large model. In this paper, we aim to investigate the potential capacity and characterize the performance of the middle layers of a generative transformer on down-stream tasks and further propose a solution that can enhance the performance of these models across all layers.

%A large language model can be defined as a sequence of transformer blocks. For example, LLAMA2 70B \cite{touvron2023llama2} has 80 transformer layers, LLAMA2 13B has 40 layers, and LLAMA2 7B has 32 layers. Currently, the common practice is to use only the contextual embeddings that has been produced by the last layer.

Large language models are revolutionizing the way we interact with information in today's world \cite{hoffmann2022training,brown2020language,refinedweb,scao2022bloom}. New models are continually emerging, demonstrating their capabilities in understanding and, more importantly, in generating human-like text. Notably, models such as ChatGPT, LLaMA 2 70B \cite{touvron2023llama2}, and Falcon 180B \cite{falcon} have had a profound impact on the applicability of large language models (LLMs). However, deploying these expansive language models can become prohibitively expensive.

What distinguishes this new era of ChatGPT-like models is their ability to perform an extraordinarily wide array of tasks in natural language processing (NLP), reasoning, and more, all through behavior cloning \cite{wei2021finetuned,wang2022self}. In fact, a single model can leverage the strong contextual learning ability offered by Standard Fine-Tuning to address numerous tasks, spanning from language comprehension to complex reasoning. While this unified usage simplifies the deployment of these models as general assistants, it remains highly inefficient. Enabling dynamic inference, where the computational resources allocated to a given query vary at inference time, can significantly enhance the practicality of employing such models in real-time scenarios. This enables the use of smaller models when the budget is limited or latency is critical. It is important to note that dynamic inference strategies for large models with a substantial number of parameters should not require loading different models during inference.

%This inefficiency arises because different tasks may require varying levels of computational resources. For instance, while a basic task like summarization can be executed adequately with a 7B model, much larger models are necessary for satisfactory code generation [cite]. Hence, dynamic inference, where the computational resources allocated to the model vary at inference time, can significantly enhance the practicality of employing such models in real-time scenarios. It's important to emphasize that, given the substantial number of parameters in these large models, a viable dynamic inference strategy should not need loading different models during inference.

Previous research has explored methods for training dynamic models capable of adapting to evolving resource constraints \cite{cai2019once,hou2020dynabert,xin2020deebert,fan2019reducing}. However, existing approaches often rely on complex training procedures or necessitate modifications to the original model architecture. SortedNet \cite{valipour2023sortednet} introduces a novel approach to training deep neural networks that leverages the inherent modularity of these networks to construct sub-models with varying computational loads. This method sorts sub-models hierarchically based on their computation/accuracy characteristics, facilitating efficient deployment during inference. Furthermore, it employs an efficient updating scheme combining random sub-model sampling with gradient accumulation to minimize the training cost. Consequently, with a single round of training, numerous models can be obtained within a single model.

While the SortedNet approach has primarily been applied to vision and language understanding tasks,  given the significant impact of generative language models in today's AI landscape, the efficacy of this method for generative tasks in NLP is of considerable interest. In fact, being able to make a large language model dynamic without the need for Pre-Training and only at the cost of a round of Standard Fine-Tuning can open doors to efficient inference of these models without incurring additional expenses associated with common model compression methods like knowledge distillation and pruning, among others.
Moreover, since all the resultant models are components of the original model, the storage requirements and the cost associated with transitioning between different computation demands become minimal. Otherwise, managing multiple models for various scenarios during inference becomes impractical.

 %In this paper, we are challenging this view as we believe the remaining layers are also powerful enough to produce good enough results. As the matter of fact, these middle layers are more efficient and their performance might be comparable for some down-stream tasks as well. Therefore, we wanted to come up with a systematic analysis that show some novel insight in this regard and guide us how we can efficiently use the power of these middle layers in addition to the last layer. 

In this study,  we challenge the conventional approach of relying solely on the last layer's contextual embeddings and use Sorted Fine-Tuning (SoFT) in place of Standard Fine-Tuning to enhance the performance of these models across multiple layers. By doing so, we aim to provide new insights into the efficiency and effectiveness of middle layers in producing high-quality results for specific downstream tasks. Our proposed approach can potentially optimize these sub-models in addition to the main model, ultimately enhancing their overall performance. In this paper, we seek to answer the following questions through systematic evaluation:

i) Do the intermediate layers resulting from Standard Fine-Tuning of a large language model generate accurate and meaningful outputs?
ii) Does Standard Fine-Tuning exhibit a sorted behavior, meaning that later layers produce more accurate and meaningful results than earlier layers? If so, to what extent? 
iii) How can we enhance this sorted behavior with minimal cost?

To answer these questions, we employ LLaMA 2 13B and perform both Standard Fine-Tuning (SFT) and Sorted Fine-Tuning (SoFT) on the Stanford Alpaca \cite{alpaca}
% GSM8K \cite{cobbe2021gsm8k} 
 and TriviaQA \cite{JoshiTriviaQA2017} datasets. For Sorted Fine-Tuning, we target 8 sub-models and share the LLM head among them to ensure cost parity. We utilize the PandaLM benchmark \cite{wang2023pandalm} to assess the performance of the sub-models on Alpaca dataset. Our findings demonstrate the superior performance of SoFT in comparison to SFT and even to memory-demanding methods like Early Exit~\cite{xin2020deebert}. %In fact with Sorted Fine-Tuning, models that are almost twice as fast as the original model can still outperform the normally tuned original model. 
The contributions of this paper can be summarized as follows:
\vspace{-2mm}
\begin{itemize}
\item Extending the SortedNet method for tuning auto-regressive language models for generative tasks by sharing a single LLM head layer among sub-models.
\vspace{-2mm}
\item Generating 8 nested sub-models, ranging from 12 to 40 layers, from LLaMA2 13B by applying Sorted Fine-Tuning on the Stanford Alpaca dataset and TriviaQA benchmarks and at a cost equivalent to Standard Fine-Tuning.
\vspace{-2mm}
\item Evaluating the performance of the sub-models of a LLaMA 2 and demonstrating the effectiveness of SoFT in enhancing the ability of intermediate layers for text generation and question answering through extensive evaluation.

\end{itemize}

\section{Related Work}
% Generative Transformers:
% LLAMA2 13B: \cite{touvron2023llama}

% Sorted Information Literature for transformers:
% SortedNet \cite{valipour2023sortednet}, DyLoRA \cite{valipour2022dyLoRA}

% Evaluation of Middle Layers in LLMs

% 70\% of attention heads and 20\% of feed-forward networks can be prunned \cite{bansal2022rethinking}

% Structural Aware Transformer language models

% Layer Dependency in Transformers

% Prunning Methods (PEFT)

%https://x.com/srush_nlp/status/1701298366087405807?s=46&t=H6t2i_XIJ0fFo_ddILR3uA

%https://arxiv.org/pdf/2307.05908.pdf

% \begin{figure*}[htb!]
% \hspace*{-1cm}
% \centering
% \resizebox{\textwidth}{!}{
%     \begin{subfigure}[b]{0.5\textwidth}
%         \resizebox{\textwidth}{!}{
%         \input{zeroshotSortedEarlyExitSFTHeatmap.tex}}
%         \label{subfig:sortedvsearlyexitsft}
%         \caption{Zero Shot Sorted vs. Early Exit SFT}
%     \end{subfigure}
%     \begin{subfigure}[b]{0.54\textwidth}
%         \resizebox{\textwidth}{!}{
%         \input{zeroShotSortedzeroShotSFTHeatmap.tex}}
%         \caption{Zero Shot Sorted vs. Zero Shot SFT}
%     \end{subfigure}
% }
% \caption{Relative comparison of Sorted LLama with respect to SFT training in zero-shot and early exit manner.}
% \label{fig:sortedvsearlyexitandother}
% \end{figure*}

\begin{table*}[hbt!]

\scriptsize
%    \begin{tabularx}{\textwidth}{|p{0.1\textwidth} | p{0.2\textwidth} | p{0.25\textwidth} | p{0.2\textwidth} | p{0.25\textwidth} |}
\resizebox{\textwidth}{!}{
    \begin{tabularx}{\textwidth}{|l | X | X |}

    \toprule
    % \multicolumn{3}{|c|}{Query:} \\ \hline
    \multicolumn{3}{|l|}{\makecell[l]{\textbf{Query}: Summarize a meeting from the given list of bullet points. Be sure to convert shorthand into a first-hand account. \\
     Rose: Analyze data and presents findings \\
     John: propose new idea \\
     Jane: appointed to head project \\
     Tom: need more time to fix software bug}} \\ \midrule %& \multicolumn{2}{l}{\makecell[l]{Define what the underlined word means for kids. \\
%\_keep a promise}} \\ \hline
    % \multicolumn{3}{|c|}{Responses:} \\ \hline
    \textbf{Layer} & \textbf{LLaMA Response (SFT+ICT)} & \textbf{Sorted LLaMA Response (SoFT)} \\ \midrule %& \textbf{SFT} & \textbf{Sorted LLaMA} \\ \hline

        12 & \color{purple}{summatronics21220f0f ...} & \color{purple}{The meeting was held in order to discuss the project and its plans for the next few months. The presenters were present in the room, and they were able to provide a detailed explanation of the project and its goals. The feedback from the meeting was positive and encouraged the team to continue working towards the goal of delivering the project.} \\ \hline %& The word is ""The word is" The 10th sentence, 10th sentence, $10th = 10th = 10$. & "A promise is a promise, no matter how small or small, always remember that you always need to go ahead and make sure that you're doing things right.!" \\ \hline

        16 & \color{purple}{The first thing we hear about these characters individually separately apart aparticularly divisidedly mannered mannerednesses ...} & \color{purple}{The meeting held at the company's headquarters, where the three named individuals were present. The first person named in the report is Rousness, who has been assigned to task task task. \color{black}{The second person named in the report is John, who has proposed a new idea. The third person named in the report is Jane,} \color{purple}{who has been assigned to task task task. The third person named in the report is Jane, who has been assigned to task task task. The second person named in the report is Jane, who has been assigned to task task task.}} \\ \hline % &  The word ""The word ""The word ""The word ""The word " \"Theramsideelteddy... & The word "keep a promise" is a simple and straightforward word that means something to a child. It means that someone is taking responsibility and is committed to delivering something to someone else. It means that someone is taking responsibility and is committed to delivering something to someone else. \\ \hline

        20 & \color{purple}{meeting held meeting held meeting held participants discussion discussion discussion notable mentionsnone none None notable mentionsnone ...} & \color{black}{The meeting held in the room with Rose, John, Jane, and Tom \color{purple}{discusses the need for analyzing data and presenting findings} \color{black}{, as well as the proposal of new ideas by John. Jane is appointed to head the project, and Tom needs more time to fix the software bug.}} \\ \hline %& The word \" ' \" means something that is kept or kept. It means that you keep promises made promises that promises that promise promises that they promises that they are not made promises that they are not made promises that they are not made. & The underlined word "kids" is an idiomatic expression that means to keep a promise or to keep something happening or happening. It means to keep things happening or happening, or to keep promises. \\ \hline

        24 & \color{purple}{meeting summary: attended attended attended attended ...} & \color{black}{The meeting was attended by Rose, John, Jane, and Tom. Rose \color{purple}{was tasked} \color{black}{with analyzing data and presenting findings. John proposed a new idea. Jane was appointed to head the project. Tom} \color{purple}{was granted} \color{black}{additional time to fix the software bug.}} \\ \hline %& When you promise something, means promise means promise means promise means promise means promise means promise means promise means promise means promise means ... & Keep a promise means to make a promise to someone, and then make sure that you do what you promised to do. It means that you are reliable and dependable. \\ \hline

        28 & \color{purple}{summary: During meeting participants discuss data analysis findings presented by Rose followed proposal new idea by John appoint Jane head project Tom request additional time fix software bug} & \color{black}{In the meeting, Rose analyzed data and presented findings, John proposed a new idea, and Jane was appointed to head the project. Tom also stated that he needed more time to fix the software bug.} \\ \hline %& When you promise something, promise means prom prom prom prom Prom ... & Keep a promise means to keep doing something or to keep a promise to someone. It means that you promise something to someone, and you do what you promise to do.\\ \hline

        32 & \color{purple}{summary: During meeting attended by Rose Analyst who analyzed data presented findings proposed new idea by John appointed Jane head project Tom Software engineer request more time fix bug} & \color{black}{In the meeting, Rose analyzed data and presented her findings, while John proposed a new idea. Jane was appointed to head the project, and Tom requested more time to fix a software bug.} \\ \hline %& When you promise something, it means you promise to do something or promise not to do something. When you keep a promise, it means you fulfill promise by doing or not doing something. For example, if you promise to clean your room, you keep promise by actually cleaning your room. promise is something you promise to do or not do, and keeping promise means fulfilling promise by doing or not doing something. & The word "keep a promise" means to do what you said you would do, even if it's difficult or takes time. It's important to keep promises because it shows others that you are trustworthy and reliable.\\ \hline

        36 & During the meeting, Rose presented her analysis of data \color{purple}{collected from various sources.} \color{black}{John proposed a new idea} \color{purple}{that could potentially improve efficiency and increase productivity.} \color{black}{Jane was appointed head of a new project} \color{purple}{that will be launched soon.} \color{black}{Tom requested more time to fix a software bug} \color{purple}{that has been affecting the company'vity since last month.} & \color{black}{In the meeting, Rose analyzed data and presented her findings, while John proposed a new idea. Jane was appointed to head the project, and Tom requested more time to fix the software bug.} \\ \hline %& Keeping a promise means promising to do something, and then actually doing it. It's like promising to clean your room, and then actually cleaning it. It's important to keep promises because it shows that you are trustworthy and reliable. & Keeping a promise means that you do what you say you will do, even if it is difficult or takes time. It means being honest and reliable, and following through on your words. For example, if you promise to clean your room, you should do it even if it is not the most fun thing to do.\\ \hline

        40 & \color{black}{At the meeting, Rose presented the analysis of the data and shared her findings with the team. John proposed a new idea} \color{purple}{that he believed could benefit the company.} \color{black}{Jane was appointed to head a new project} \color{purple}{that the company had recently taken on.} \color{black}{Tom, however, requested more time to fix a software bug that had been discovered.} \color{purple}{The team discussed the importance of the bug fix and agreed to give Tom the additional time he needed.} & \color{black}{In the meeting, Rose analyzed data and presented her findings. John proposed a new idea, and Jane was appointed to head the project. Tom requested more time to fix the software bug.} %& To keep a promise means to do what you said you would do, even when it'ioned hard or you don't feel like it. It's like keeping your word, and it's important to keep your promises because it shows that you're trustworthy and that people can count on you. & The underlined word "keep a promise" means to do what you say you will do, even if it's hard or takes a long time. For example, if you promise your friend to play with them after school, you should keep that promise and play with them after school. \\ \hline

        % \multicolumn{3}{|c|}{Query:} \\ \hline
        \\ \bottomrule
    \end{tabularx}}
    \caption{\small{Generated responses in different layers of LLaMA (Early Exit) and Sorted LLaMA for two samples from PandaLM validation set. \color{black}{Black} \color{black}{phrases mean the answer is following the instruction and} \color{purple}{Red} \color{black}{ones demonstrate errors and hallucinations.}}}
    \label{tab:examples}
\end{table*}

This section briefly introduces the most relevant papers to our work. 
\paragraph{Many-in-One Models}

\begin{figure*}[!htb]
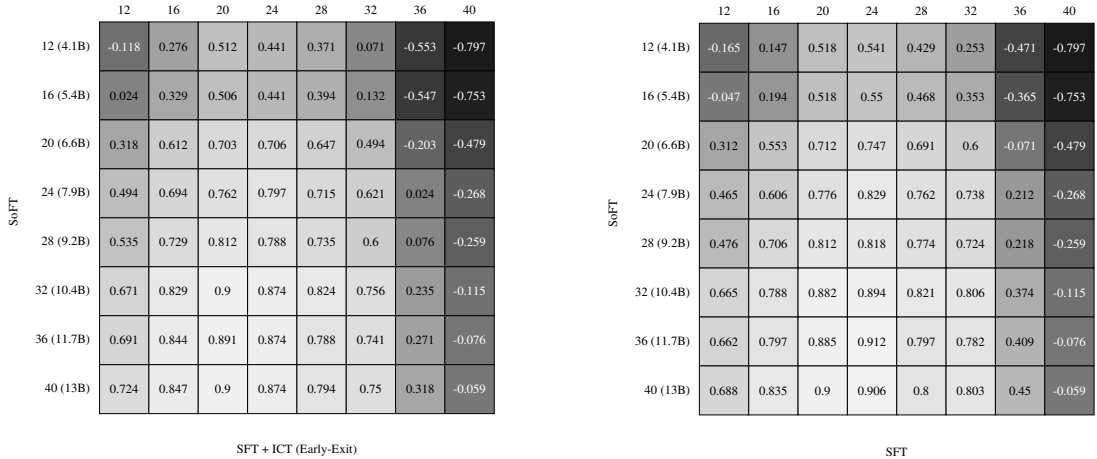

    % [hbt!]
    \hspace*{-1cm}
    \centering
    
    \resizebox{\columnwidth}{!}{
    \begin{subfigure}[t]{1.1\textwidth}
            \input{zeroshotSortedEarlyExitSFTHeatmap_v2}
            \label{subfig:zeroshotsortedearlyexitsft}
    \end{subfigure}
    }%
    ~
    \resizebox{\columnwidth}{!}{
    \begin{subfigure}[t]{1.1\textwidth}
            \input{zeroShotSortedzeroShotSFTHeatmap_v2}
            \label{subfig:zeroshotsortedezeroshotsft}
    \end{subfigure}
    }
    \caption{\small{SoFT vs. SFT + ICT (Early-Exit) (Left) and SoFT vs. SFT (Right). Note that for our SoFT method, the output prediction layer is shared between all sub-models whereas, for Early-Exit, a separate prediction head is learned per sub-model, making inference inefficient. Both SoFT and SFT had equivalent training time (2 Epochs) in this experiment. The number in each cell is calculated by considering wins as the times SoFT sub-models (rows) were preferred, losses as the times SFT sub-models (columns) were preferred and ties when non of them were preferred (Equation \ref{eq:score}).} Algorithm performance is correlated to cell whiteness: white is better, zero is on-par, dark is worse.}
    \label{fig:results}
\end{figure*}

Deep neural networks (DNNs) are often overparameterized, motivating researchers to explore ways to use the parameters of the models more efficiently. More number of parameters lead to higher costs of deployment for neural networks. Moreover, in practice, these overparametrized DNNs are expected to accommodate customers with varying requirements and computational resources. To address these diverse demands, one can think of training models of different sizes, which can be prohibitively costly (in terms of training and memory), or another alternative is to train many-in-one networks~\cite{cai2019once}. Many-in-one solutions aim to train a network along with some of its sub-networks simultaneously for specific tasks. For example, we can consider the \textit{Early-Exit} method~\cite{xin2020deebert}, wherein a prediction head is fine-tuned on top of specific intermediate layers within a network. 
Another approach is \textit{Layer Drop}~\cite{fan2019reducing}, which trains a network in any depth by randomly dropping the layers during training. While both Early-Exit and Layer Drop are simple solutions, they are not state-of-the-art in terms of performance. In Early-Exit, we only train the output prediction layer on top of each intermediate layer, and this layer might not have enough capacity to retain a good performance. Layer Drop, conversely, suffers from the abundant number of possible sub-models in training, which makes the training process exhaustive and sub-optimal. Furthermore, this approach requires tuning the extent of dropping layers during training. This additional hyper-parameter, layer drop rate during training determines the best size and setting of the model at the inference time. Deviating from the training drop rate at the inference time can result in a significant drop in performance. 

~\citet{cai2019once} in \textit{Once for All (OFA)} proposed an alternative solution to neural architecture search (NAS). OFA requires training the model and all possible sub-models in an arbitrary progressive way followed by a separate search phase. 
Dyna-BERT~\cite{hou2020dynabert} is another work that targets training Dynamic pre-trained many-in-one BERT models in two stages: first, distilling from the main network to the width adaptive networks and then distilling from the width adaptive networks to depth adaptive networks. Both width adaptive and depth adaptive networks have a limited pre-defined set of width and depth for the sub-models. While both OFA and DynaBERT have shown successful results, their solutions are hardly applicable to multi-billion-parameter LLMs because of their complicated multi-stage training process and their search and knowledge distillation requirements.  
SortedNet~\cite{valipour2023sortednet} is a recent method that forms and trains sub-models of a network in a sorted manner while not requiring any search during training or inference. SortedNet has shown superior performance compared to other previously mentioned methods in terms of simplicity, performance, scalability, and generalization. Considering these benefits, we target deploying the SortedNet training algorithm for developing many-in-one LLMs.

\paragraph{Many-in-One Large Language Models (LLMs)} 
Large language models have recently gained significant attention in the literature ~\cite{touvron2023llama,brown2020language,openai2023gpt4,chowdhery2022palm,ouyang2022training}. In practice, these LLMs serve users with different tasks, expectations, and computational budget requirements~\cite{sun2022black}. There are two types of adaptation approaches to make LLMs suitable for customer requirements: first is the so-called parameter efficient tuning (PEFT), and second is model compression. In PEFT, the core backbone model remains the same, and we just update much smaller adapter parameters (e.g. LoRA~\cite{hu2021LoRA}, KRONA~\cite{edalati2022krona}, Adapter~\cite{houlsby2019parameter,pfeiffer2020adapterfusion}, DyLoRA~\cite{valipour2022dyLoRA}, Ladder Side-Tuning~\cite{sung2022lst}) and Compacter~\cite{karimi2021compacter}. In model compression, the larger model is compressed using any model compression solutions such as knowledge distillation \cite{hinton2015distilling, hsieh2023distilling, wu2023lamini}, pruning~\cite{bansal-etal-2023-rethinking}, and quantization~\cite{prato2019fully, dettmers2023qlora}, a good related survey can be found in~\cite{zhu2023survey}. Even though PEFT solutions are pretty popular with LLMs, they do not provide dynamic-size LLMs. Model compression solutions can provide models with different sizes, but they need to train each compressed model separately, and they are not many-in-one models.

To the best of our knowledge, this work is the first attempt to obtain many-in-one generative LLMs by applying the Sorted Fine-Tuning to the LLaMA 13B model. 
Considering the benefits of many-in-one networks and the growing application of LLMs, we are hoping that this will help the community to build more efficient large language models that can be deployed adaptive during inference using methods such as SoFT.

\section{Methodology}
\label{sec:methodology}

This paper focuses on making generative LLMs many-in-one by unlocking the potential of intermediate layers through the SortedNet approach~\cite{valipour2023sortednet}.

%The methodology of this paper concerns making LLMs many-in-one inspired by the SortedNet approach~\cite{valipour2023sortednet}. 
Let's consider a language model $f(x;\theta)$ with the parameters $\theta$ and the input $x$. The following is the sorted training procedure: 

\paragraph{Forming Sub-Networks} First, we need to form the sub-networks of the LLM. For the sake of simplicity and without loss of generality, we focus on the depth-wise sub-networks. Supposed that the sub-network $f_n(x;\theta_n)$ refers to the first $n$ layers of $f(x;\theta)$. In this paper, the language model is considered to be LLaMA2 13B. Since LLaMA2 comprises 40 layers, we define the sub-networks as $n\in \mathbf{B}=\{ 12, 16, 20, 24, 28, 32, 36, 40 \}$.  

\paragraph{Calculating the Output of Sub-Networks}

Each sub-model's output will be predicted using the shared output prediction head from the last layer (original network). Remember that in the LLaMA model, there is an RMSNorm layer \cite{zhang2019root} before the output prediction head. This RMSNorm is added before the shared prediction head of every sub-model. This normalization may be an important factor that helps Sorted LLaMA to generalize better for all sub-models.
\vspace{-2mm}
\paragraph{Objective Function}
Let $L_n(x;\theta_n)$ be the loss for the $n^\text{th}$ sub-model for input batch $x$. To train the network, we define the loss as the summation of the losses of all these sub-models:
\vspace{-1mm}
\begin{equation}
    \begin{split}
        & \mathcal{L} = \dfrac{\sum_{n\in \mathbf{B}} L_n (x;\theta_n)}{\mathbf{|B|}}   %\\ 
        %& \theta^+ = \theta^- -\eta \nabla_{\theta} \mathcal{L}
    \end{split}
\end{equation}

For the experiments conducted in the paper, $|\mathbf{B}| = 8$. Note that these sub-models have shared parameters through a nested style i.e. $\theta_1 \subset \theta_2 ... \subset \theta_n$.

% For each sample, we apply a sequence of n transformer blocks $f(x;\theta(i))|^{n}_{i=1}$, where $\theta(i)$ represents the weight of the $i^{th}$ block, to get the ultimate contextual embeddings of an LLM. Before passing the learned contextual embedding to a language model head to find the probability distribution over language tokens, we use the more efficient RMSNorm \cite{zhang2019root} to normalize the output representation. We believe this normalization is a key factor that helps SortedNet to generalize better for all sub-models. 

%TODO: run an ablation study that supports this claim

% We have added the RMSNorm before the intermediate output classifiers. 

\paragraph{Training Dataset}
% \subparagraph{Stanford Alpaca}

We utilized the Stanford Alpaca dataset \cite{alpaca}, which includes demonstrations of 52K instruction-following examples.
% and GSM8K benchmark \cite{cobbe2021gsm8k}, which consists of 8.5K mathematical reasoning questions.
We also used TriviaQA open-domain QA benchmark \cite{JoshiTriviaQA2017} including 110K closed-book question-answer pairs. % TODO: Why we choose this data? 
%TODO: Some details for the collection maybe

% \paragraph{Normalization}
% \subparagraph{RMSNorm}
\vspace{-2mm}
\paragraph{Evaluation}

In this paper, in addition to embedding the last layer, we evaluate the quality of the embeddings of intermediate outputs spanning from block 1 to n. PandaLM benchmark \cite{wang2023pandalm} compares the output of different sub-models. PandaLM deploys a large language model (Fine-Tuned LLaMA 7b) to judge the quality of generated text from two sources. PandaLM provides a validation set consisting of 170 instructions\footnote{github.com/WeOpenML/PandaLM/blob/main/data/testset-inference-v1.json}, to evaluate target models for instruction-following tasks. To ensure that the order of the models' responses does not influence the judgment of the PandaLM evaluator, we reported an average score under both the Model 1 first and the Model 2 first scenarios. The output of the PandaLM evaluation is the number of wins, denoted as $W$, the number of losses, denoted as $L$, and the number of ties, denoted as $T$, in the validation set. The final reported score has been calculated using the following formula:

\begin{equation}
\label{eq:score}
    Score = \dfrac{(W - L)}{T}
\end{equation}

The final score is a number between -1 and 1, in which 1 represents a strong win rate and -1 means a poor performance of the model.

We used accuracy (exact match) as the evaluation metric for the TriviaQA benchmark.

\begin{figure}[htbp]
\resizebox{0.49\columnwidth}{!}{
    \begin{minipage}[t]{1.15\columnwidth}
        %The matrix in numbers
%Horizontal target class
%Vertical output class
\def\myConfMat{{{-0.05, -0.556, -0.668, -0.756}, {0.068, -0.468, -0.609, -0.721}, {0.385, -0.168, -0.385, -0.503}, {0.506, 0.053, -0.156, -0.259}, {0.582, 0.071, -0.085, -0.212}, {0.721, 0.321, 0.112, -0.068}, {0.697, 0.341, 0.159, -0.056}, {0.668, 0.382, 0.194, -0.041}}}

\def\winRateMat{{{83.0, 40.0, 29.5, 19.0}, 
{78.0, 33.0, 23.0, 12.5}, 
{111.5, 58.5, 46.5, 30.0}, 
{130.0, 77.0, 52.0, 37.0}, 
{130.5, 73.5, 51.5, 44.0}, 
{140.5, 85.0, 62.0, 42.0}, 
{141.0, 96.0, 75.5, 48.5}, 
{145.5, 105.0, 80.5, 56.0}}}

\def\loseRateMat{{{84.5, 128.0, 137.0, 147.0}, 
{89.5, 131.5, 144.0, 151.5}, 
{51.0, 84.5, 104.5, 119.0}, 
{33.0, 66.5, 87.5, 100.5}, 
{33.5, 72.5, 82.5, 91.5}, 
{23.0, 58.5, 72.5, 85.0}, 
{23.0, 47.0, 57.5, 68.5}, 
{18.0, 39.0, 50.0, 62.0}}}

\def\tieRateMat{{{2.5, 2.0, 3.5, 4.0}, 
{2.5, 5.5, 3.0, 6.0}, 
{7.5, 27.0, 19.0, 21.0}, 
{7.0, 26.5, 30.5, 32.5}, 
{6.0, 24.0, 36.0, 34.5}, 
{6.5, 26.5, 35.5, 43.0}, 
{6.0, 27.0, 37.0, 53.0}, 
{6.5, 26.0, 39.5, 52.0}}}

\def\xNames{{"12","20","28","36"}}

\def\yNames{{"12 (4.1B)","16 (5.4B)","20 (6.6B)","24 (7.9B)","28 (9.2B)","32 (10.4B)", "36 (11.7B)", "40 (13B)"}}

\def\numX{4}
\def\numY{8}

\def\myScale{1.5} % 1.5 is a good scale. Values under 1 may need smaller fonts!

\begin{tikzpicture}[
    scale = \myScale,
    %font={\scriptsize}, %for smaller scales, even \tiny may be useful
    font={\normalsize},
    ]

\tikzset{vertical label/.style={rotate=90,anchor=east}}   % usable styles for below
\tikzset{diagonal label/.style={rotate=45,anchor=north east}}

\foreach \y in {1,...,\numY} %loop vertical starting on top
{
    % Add class name on the left
    \node [anchor=east] at (0.4,-\y) {\pgfmathparse{\yNames[\y-1]}\pgfmathresult}; 

    % Add class name on the left
    %\node [anchor=south] at (0.4,-\x) {\pgfmathparse{\xNames[\x-1]}\pgfmathresult};
    
    \foreach \x in {1,...,\numX}  %loop horizontal starting on left
    {
% %---- Start of automatic calculation of totSamples for the column ------------   
%     \def\totSamples{0}
%     \foreach \ll in {1,...,\numClasses}
%     {
%         \pgfmathparse{\myConfMat[\ll-1][\x-1]}   %fetch next element
%         \xdef\totSamples{\totSamples+\pgfmathresult} %accumulate it with previous sum
%         %must use \xdef fro global effect otherwise lost in foreach loop!
%     }
%     \pgfmathparse{\totSamples} \xdef\totSamples{\pgfmathresult}  % put the final sum in variable
% %---- End of automatic calculation of totSamples ----------------
    
    \begin{scope}[shift={(\x,-\y)}]
        \def\mVal{\myConfMat[\y-1][\x-1]}

        \def\wVal{\winRateMat[\y-1][\x-1]}
        \def\lVal{\loseRateMat[\y-1][\x-1]}
        \def\tVal{\tieRateMat[\y-1][\x-1]}
        
        % \def\scoreVal{\myWinRateMat[\y-1][\x-1]}
        % The value at index y,x (-1 because of zero indexing)
        \pgfmathsetmacro{\r}{\mVal}   %

        \pgfmathsetmacro{\w}{\wVal}
        \pgfmathsetmacro{\l}{\lVal}
        \pgfmathsetmacro{\t}{\tVal}
        
        %\pgfmathsetmacro{\s}{\scoreVal}  
        \pgfmathtruncatemacro{\p}{\r}
        \pgfmathtruncatemacro{\c}{(\r+1)/2*100}
        \coordinate (C) at (0,0);
        \ifthenelse{\c<50}{\def\txtcol{white}}{\def\txtcol{black}} %decide text color for contrast
        \node[
            draw,                 %draw lines
            text=\txtcol,         %text color (automatic for better contrast)
            align=center,         %align text inside cells (also for wrapping)
            fill=white!\c!black,        %intensity of fill (can change base color)
            minimum size=\myScale*10mm,    %cell size to fit the scale and integer dimensions (in cm)
            inner sep=0,          %remove all inner gaps to save space in small scales
            ] (C) {\r};     %text to put in cell (adapt at will)
        %Now if last vertical class add its label at the bottom
        
        \ifthenelse{\y=\numY}{
        \node [] at ($(C)+(0,7.75)$) % can use vertical or diagonal label as option
        {\pgfmathparse{\xNames[\x-1]}\pgfmathresult};}{}
    \end{scope}
    }
}
%Now add x and y labels on suitable coordinates
\coordinate (yaxis) at (-1.25,0.0-\numY/2);  %must adapt if class labels are wider!
\coordinate (xaxis) at (0.5+\numX/2, -\numY-1.25); %id. for non horizontal labels!
\node [vertical label] at (yaxis) {SoFT};
\node []               at (xaxis) {Extracted Fine-Tuning};
\end{tikzpicture}
        % \caption{caption1}
        \label{subfig:zeroshotcapacity2vs2}
    \end{minipage}}%
    \hfill%
\resizebox{0.49\columnwidth}{!}{
    \begin{minipage}[t]{1.15\columnwidth}
        %The matrix in numbers
%Horizontal target class
%Vertical output class
\def\myConfMat{{{0.138, -0.453, -0.55, -0.659}, {0.265, -0.276, -0.35, -0.524}, {0.565, 0.032, -0.156, -0.291}, {0.597, 0.226, 0.044, -0.171}, {0.685, 0.226, 0.038, -0.171}, {0.741, 0.403, 0.15, -0.038}, {0.756, 0.418, 0.235, 0.044}, {0.788, 0.397, 0.271, 0.053}}}

\def\winRateMat{{{99.5, 50.5, 40.0, 27.0}, 
{113.5, 59.5, 48.5, 36.5}, 
{127.5, 70.0, 54.5, 36.0}, 
{137.0, 90.5, 70.5, 50.0}, 
{134.5, 92.0, 76.0, 55.0}, 
{143.5, 102.5, 77.0, 59.0}, 
{148.0, 107.0, 87.0, 61.0}, 
{146.0, 107.0, 94.0, 69.0}}}

\def\loseRateMat{{{67.5, 115.0, 126.0, 136.0}, 
{52.0, 100.0, 114.5, 125.0}, 
{34.0, 71.0, 90.5, 104.0}, 
{24.0, 52.5, 68.5, 86.5}, 
{27.5, 51.0, 61.0, 78.5}, 
{20.0, 41.5, 58.0, 66.5}, 
{15.5, 41.0, 49.0, 64.5}, 
{17.5, 39.5, 48.0, 54.0}}}

\def\tieRateMat{{{3.0, 4.5, 4.0, 7.0}, 
{4.5, 10.5, 7.0, 8.5}, 
{8.5, 29.0, 25.0, 30.0}, 
{9.0, 27.0, 31.0, 33.5}, 
{8.0, 27.0, 33.0, 36.5}, 
{6.5, 26.0, 35.0, 44.5}, 
{6.5, 22.0, 34.0, 44.5}, 
{6.5, 23.5, 28.0, 47.0}}}

\def\xNames{{"12","20","28","36"}}

\def\yNames{{"12 (4.1B)","16 (5.4B)","20 (6.6B)","24 (7.9B)","28 (9.2B)","32 (10.4B)", "36 (11.7B)", "40 (13B)"}}

\def\numX{4}
\def\numY{8}

\def\myScale{1.5} % 1.5 is a good scale. Values under 1 may need smaller fonts!

\begin{tikzpicture}[
    scale = \myScale,
    %font={\scriptsize}, %for smaller scales, even \tiny may be useful
    font={\normalsize},
    ]

\tikzset{vertical label/.style={rotate=90,anchor=east}}   % usable styles for below
\tikzset{diagonal label/.style={rotate=45,anchor=north east}}

\foreach \y in {1,...,\numY} %loop vertical starting on top
{
    % Add class name on the left
    \node [anchor=east] at (0.4,-\y) {\pgfmathparse{\yNames[\y-1]}\pgfmathresult}; 

    % Add class name on the left
    %\node [anchor=south] at (0.4,-\x) {\pgfmathparse{\xNames[\x-1]}\pgfmathresult};
    
    \foreach \x in {1,...,\numX}  %loop horizontal starting on left
    {
% %---- Start of automatic calculation of totSamples for the column ------------   
%     \def\totSamples{0}
%     \foreach \ll in {1,...,\numClasses}
%     {
%         \pgfmathparse{\myConfMat[\ll-1][\x-1]}   %fetch next element
%         \xdef\totSamples{\totSamples+\pgfmathresult} %accumulate it with previous sum
%         %must use \xdef fro global effect otherwise lost in foreach loop!
%     }
%     \pgfmathparse{\totSamples} \xdef\totSamples{\pgfmathresult}  % put the final sum in variable
% %---- End of automatic calculation of totSamples ----------------
    
    \begin{scope}[shift={(\x,-\y)}]
        \def\mVal{\myConfMat[\y-1][\x-1]}

        \def\wVal{\winRateMat[\y-1][\x-1]}
        \def\lVal{\loseRateMat[\y-1][\x-1]}
        \def\tVal{\tieRateMat[\y-1][\x-1]}
        
        % \def\scoreVal{\myWinRateMat[\y-1][\x-1]}
        % The value at index y,x (-1 because of zero indexing)
        \pgfmathsetmacro{\r}{\mVal}   %

        \pgfmathsetmacro{\w}{\wVal}
        \pgfmathsetmacro{\l}{\lVal}
        \pgfmathsetmacro{\t}{\tVal}
        
        %\pgfmathsetmacro{\s}{\scoreVal}  
        \pgfmathtruncatemacro{\p}{\r}
        \pgfmathtruncatemacro{\c}{(\r+1)/2*100}
        \coordinate (C) at (0,0);
        \ifthenelse{\c<50}{\def\txtcol{white}}{\def\txtcol{black}} %decide text color for contrast
        \node[
            draw,                 %draw lines
            text=\txtcol,         %text color (automatic for better contrast)
            align=center,         %align text inside cells (also for wrapping)
            fill=white!\c!black,        %intensity of fill (can change base color)
            minimum size=\myScale*10mm,    %cell size to fit the scale and integer dimensions (in cm)
            inner sep=0,          %remove all inner gaps to save space in small scales
            ] (C) {\r};     %text to put in cell (adapt at will)
        %Now if last vertical class add its label at the bottom
        
        \ifthenelse{\y=\numY}{
        \node [] at ($(C)+(0,7.75)$) % can use vertical or diagonal label as option
        {\pgfmathparse{\xNames[\x-1]}\pgfmathresult};}{}
    \end{scope}
    }
}
%Now add x and y labels on suitable coordinates
\coordinate (yaxis) at (-1.25,0.0-\numY/2);  %must adapt if class labels are wider!
\coordinate (xaxis) at (0.5+\numX/2, -\numY-1.25); %id. for non horizontal labels!
\node [vertical label] at (yaxis) {SoFT};
\node []               at (xaxis) {Extracted Fine-Tuning};
\end{tikzpicture}
        % \caption{caption2}
        \label{subfig:zeroshotcapacity4vs2}
    \end{minipage} 
}
\caption{\small{SoFT vs. Extracted Fine-Tuning. The left figure shows an equal training time setup (2 epochs), and the figure on the right considers two extra training epochs for SoFT.} 
% \moji{Algorithm performance is correlated to cell whiteness: white is better, zero is on-par, negative is worse.}
}
\label{fig:SoFTcapacity}
\end{figure}

\paragraph{Baseline}

The primary objective of the LLM in this paper is to follow the provided instructions by a query. 
% \textcolor{red}{do we need to add GSM8K}. 
Therefore, following the setup of Alpaca \cite{alpaca}, we fine-tuned LLaMA2 13B on the Stanford Alpaca Dataset with two setups: (1) Regular Standard Fine-Tuning (SFT) as the baseline, focusing only on the training of the last layer of the network as the common practice in the literature; (2) Sorted Fine-Tuning (SoFT), calculating loss for multiple outputs from layer 12 to layer 40 (last layer) with four intervals, and training multiple models simultaneously as explained in the previous section.

% \subparagraph{LLAMA2-SFT:} A supervised fine-tuned version of LLAMA2 13B

% \subparagraph{LLAMA2-CHAT:} An instruction-tuned LLAMA2 13B model

% \paragraph{Sorted-LLAMA2}

% \paragraph{Evaluation}
% \subparagraph{Panda-LM \cite{wang2023pandalm}}

\begin{figure}[!t]
    \centering
    \scalebox{0.19}{\includegraphics{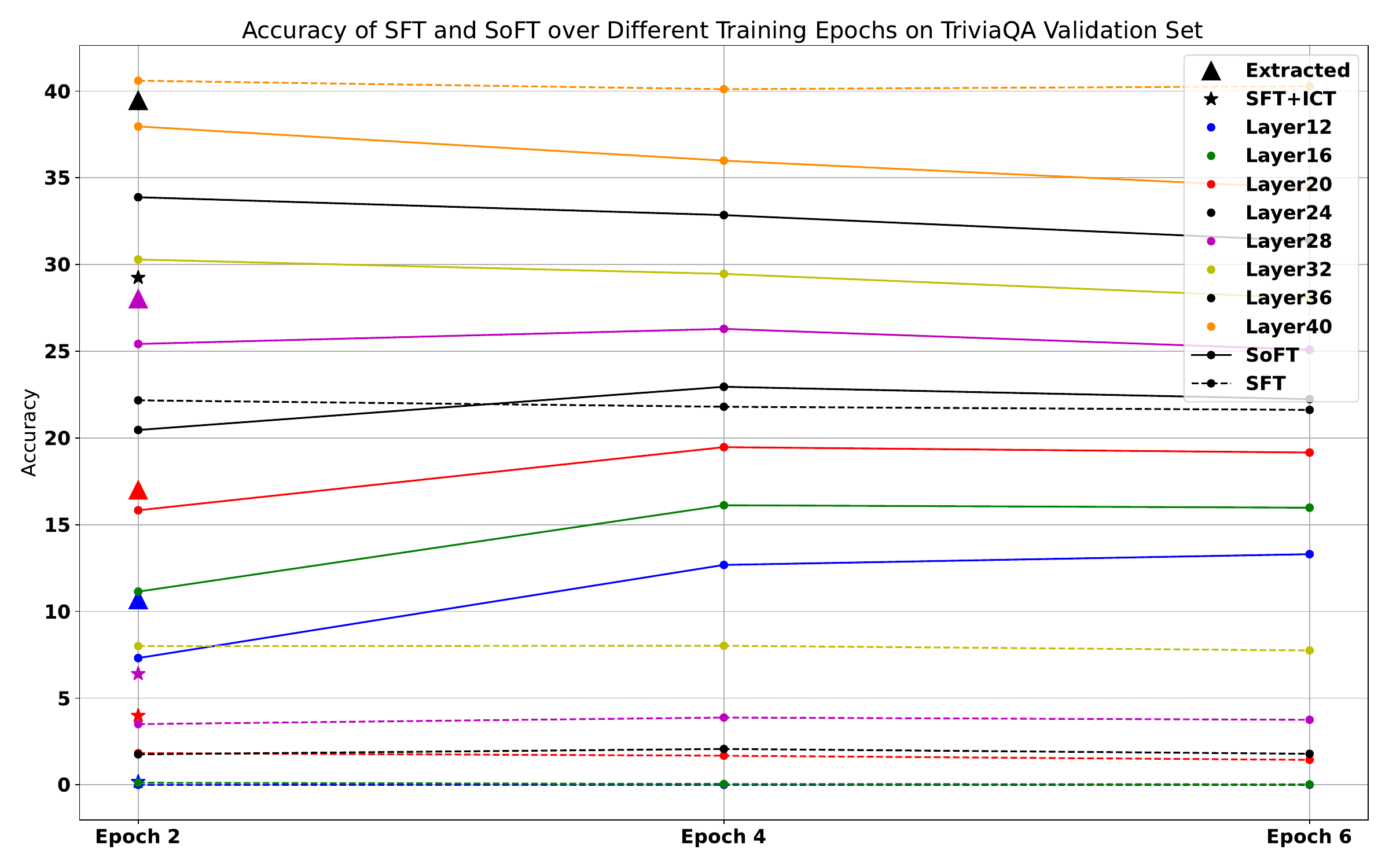}}
    \caption{\small{The results of TriviaQA. We reported case-sensitive exact match accuracy as the main metric. SFT+ICT and Extracted Fine-Tuned results can be found in Epochs 2, as we found Epoch 2 checkpoint saturated for the original SFT experiment (main LLaMA2 13b model with 40 layers).}
    \vspace{-3mm}
    }
    \label{fig:triviaqa}
\end{figure}

\section{Experiments}
\label{sec:experiments}
%\subsection{Sorted LLaMA Performance:}

This section delves into the experiments' specifics and the analysis provided to understand better the effect of Sorted Fine-Tuning over the performance of a large language model like LLaMA2 \cite{touvron2023llama2}. 
The details of the experimental setup used for these experiments are available in the appendix \ref{appx:experimental_setup}.
Before diving into results, we are going to define certain notations that we used for different setups in our experiments:
\vspace{-2mm}
\begin{itemize}
    \item \textbf{SoFT/SFT}: We first train the model with SoFT or SFT paradigms and use the sub-models after training without any further training of the language model head for intermediate layers.
    % \item \textbf{Intermediate Classifier Tuning (ICT):} It refers to the scenario where we further tune the classifier heads of the intermediate layers by training for more epochs.
\vspace{-3mm}
    \item \textbf{SFT+Intermediate Classifier Tuning (ICT)}: We first train the model with SFT paradigm and then further fine-tune the language model head exclusively for each sub-model while keeping their weights frozen. The SFT+ICT is also known as Early-Exit \cite{xin2020deebert} in the literature.
\vspace{-3mm}
    \item \textbf{Extracted Fine-Tuning}: We extract the sub-models from the learned weights of the pre-trained original model and train each sub-model separately. 
    % \item \textbf{SFT+Extracted Fine-Tuning}: When we extract the sub-models from the learned SFT weights and train each sub-model separately. 
\end{itemize}

% \subsection{Setups}
%for both SFT as the baseline and SoFT as the desired model.

\begin{table*}[t]
 \setlength\extrarowheight{2pt}
 \centering
 \scalebox{0.7}{\begin{tabular}{l|ccc|ccc}
  
  % \cline{1-3}
  \toprule
   \multicolumn{4}{c}{\textbf{PandaLM}} & \multicolumn{3}{c}{\textbf{TriviaQA}} 
   % & \multicolumn{3}{c}{\textbf{GSM8K}} 
   \\ \midrule
   \multicolumn{7}{c}{\textbf{Auto-regressive Decoding}} \\
   \textbf{Model} & \textbf{Time per Token (ms)} & \textbf{Score} & \textbf{Rejection Ratio} & \textbf{Time per Token (ms)} & \textbf{Accuracy} & \textbf{Rejection Ratio} 
   % & \textbf{Time per Token (ms)} & \textbf{Accuracy} & \textbf{Rejection Ratio}
   \\ 
   Layer 40 (full) & 94.07 & - & - & 91.27 & 37.95 & - 
   % & 93.60 & 33.05 & -
   \\
   \midrule
   \multicolumn{7}{c}{\textbf{Speculative Decoding}} \\
   \textbf{Draft Model} & \textbf{Time per Token (ms)} & \textbf{Score} & \textbf{Rejection Ratio} & \textbf{Time per Token (ms)} & \textbf{Accuracy} & \textbf{Rejection Ratio}
   % & \textbf{Time per Token (ms)} & \textbf{Accuracy} & \textbf{Rejection Ratio}
   \\ %\midrule 
   Layer 12 & 80.86 (1.16$\times$) & -0.144 & 0.37 & 110.50 (0.82$\times$) & 34.36 & 0.72 
   % & 66.10 (1.41$\times$) & 32.22 & 0.43 
   \\
   Layer 16 & 84.10 (1.11$\times$) & -0.211 & 0.31 & 118.92 (0.76$\times$) & 34.16 & 0.70
   % & 67.51 (1.38$\times$) & 33.20 & 0.32
   \\ 
   Layer 20 & 84.50 (1.11$\times$) & -0.144 & 0.26 & 139.78 (0.65$\times$) & 34.19 & 0.66 
   % & 68.45 (1.36$\times$) & 33.73 & 0.23
   \\ \midrule
   \multicolumn{7}{c}{\textbf{Instance-Aware Dynamic Inference}}\\
   \textbf{Model} & \textbf{Time per Token (ms)} & \textbf{Score} & \textbf{Rejection Ratio} & \textbf{Time per Token (ms)} & \textbf{Accuracy} & \textbf{Rejection Ratio}
   % & \textbf{Time per Token (ms)} & \textbf{Accuracy} & \textbf{Rejection Ratio}
   \\ %\midrule 
   Layer 12:40 & 69.91 (1.34$\times$) & -0.050 & - & 81.01 (1.12$\times$) & 36.53 & - 
   % & 51.30 (1.82$\times$) & 30.62 & -
   \\
   \bottomrule
 \end{tabular}}
 % \vspace{-3mm}
 \caption{
  \small{Speed-up in inference time on three PandaLM and TriviaQA benchmarks by utilizing Speculative Decoding and Instance-Aware Dynamic Inference techniques. Score column in PandaLM section means the score of the model versus the Auto-regressive generated results based on Equation \ref{eq:score}.}
  % \vspace{-4mm}
 }
 \label{tab:speculative}
\end{table*}

\subsection{What is the effect of sorting information across layers of a generative model?}

As mentioned before, we generated responses for all the layers $n\in \mathbf{B}$ for both SFT and SoFT-based trained models. Then, we conducted a pair-wise comparison between all the sub-models in the two trained models using the PandaLM evaluator. As the results suggest in Figure \ref{fig:results}, sorted training significantly unlocks the potential of intermediate layers in generating the desired output. Some generated examples can be found in Table \ref{tab:examples}.

Sorted LLaMA (aka SoFT) is outperforming regular fine-tuning (SFT) in nearly all layer comparisons by a meaningful margin, as shown through automated evaluation in Figure \ref{fig:results}. 

It might be noted that the Layer 12 performance of SFT is slightly better compared to Layer 12 of Sorted LLaMA. We argue this is happening because the outputs of early layers in SFT are mostly gibberish (see Table \ref{tab:examples} as an example), and the PandaLM evaluator has not been trained on such data. Hence the automatic evaluation results for this layer are not meaningful. To further investigate the reason behind the results for early sub-models, we conducted human evaluation on 6 cells of two tables in Figure \ref{fig:results} (Layer 12 of SFT and SFT+ICT vs Layers 12,16, and 20 SoFT) to verify our claim. We observed that SoFT early sub-models could significantly outperform sub-model layer 12 of both SFT and SFT+ICT models, proving the negative impact of gibberish text on PandaLM evaluator performance. As we go to higher layers in SFT, the generated text becomes meaningful, which makes the comparison with the Sorted LLaMA layer counterpart more reasonable. 

%TODO: seems to have overlapping information with the notation section SFT+ICT
Moreover, to improve SFT results, inspired by Early-Exit \cite{xin2020deebert}, we also tried the scenario in which a separate classifier head is dedicated to all sub-models of SFT. This method has been introduced in the notation section as SFT+ICT. These classification heads have been trained an additional epoch after SFT tuning while keeping the base model frozen. Note that this setting suffers from significant memory overhead during tuning and inference compared to our SoFT method. In fact, the extra number of parameters for SFT+ICT (Early Exit) is $|B|-1 \times D \times V$, where $|B|$ is the number of sub-models, $D$ is the hidden size of the model, and $V$ is the vocabulary size. For LLaMA 2 13B, this is equivalent to 1B extra parameters.  

The results of comparing sorted with the early exit are shown in figure \ref{fig:results} (Left). Despite having far more parameters, SFT+ICT (Early-Exit) underperforms our sorted tuning for most sub-models. 
%Also, another area in which the SFT model performs better compared to Sorted LLaMA is in the last two columns, where different layers of Sorted LLaMA are compared to the last two sub-models of the SFT model. The fine-tuned model performs better only with a larger number of parameters. 
%For example, 32 layers in Sorted LLaMA won't beat the 40 layers SFT.
According to the results, the sub-model in Sorted LLaMA with 36 layers performs almost as well as regular fine-tuning of the full-size model. This showcases the impressive ability of our proposed paradigm to generate powerful, small sub-models that perform similarly to the original model.
% In addition, we analyzed the effect of training different classifiers, SoFT+ICT, in the appendix \ref{appx:additional_experiments}.
Another experiment that has been conducted in appendix \ref{appx:additional_experiments}, further investigated the impact of longer training time for SoFT. The results show that our model was still under-trained, and we could observe a significant improvement in Sorted LLaMA performance with longer training time.

% We also analyzed the effect of early exit on SoFT and another scenario that considers more training time for SoFT and observed a significant improvement in Sorted LLaMA performance. 
% (Table \ref{tab:pandalm_results_sft_SoFT_4epochs}). 
Moreover, we compared the performance of Sorted LLaMA sub-models with the actual capacity of these models by fine-tuning the sub-models separately and reporting the results in both equal training time and more training time for SoFT. 
% \st{(see \ref{appx:capacity} and Table \ref{tab:pandalm_results_middleLayer_sorted})}.
We extracted 4 sub-models (Layer 12, Layer 20, Layer 28, and Layer 36) and each time fully fine-tuned the extracted sub-model separately for two epochs on the Alpaca dataset. Figure \ref{fig:SoFTcapacity} and Table \ref{tab:pandalm_results_middleLayer_sorted} shows the comparison between Extracted Fine-Tuned and SoFT sub-models. The first part in Table \ref{tab:pandalm_results_middleLayer_sorted} shows the equal training budget setup (2 Epochs) comparison in which SFT demonstrates slightly better performance compared to the similar SoFT sub-models. Further training SoFT will lead to better sorted sub-models in which SoFT outperforms the fully fine-tuned sub-models, proving the positive impact of SoFT on the performance of lower sub-models. 

The analysis of the computational overhead of SoFT can be found in \ref{appx:comp_overhead}.
% Previously, \cite{} \textcolor{red}{cite here} showed that extended knowledge distillation can happen by transferring information from upper layer to lower ones.

%In figure \ref{fig:earlyexitearlyexit}, we evaluated both SFT and SoFT in an equal Early-Exit framework. As depicted, the outcome remains almost unchanged.

% \begin{table}[t]
%  \setlength\extrarowheight{2pt}
%  \centering
%  \scalebox{0.65}{\begin{tabular}{l|cccc}
  
%   % \cline{1-3}
%   \toprule
%    \textbf{Sub-model (layer)} & \textbf{SFT} & \textbf{SoFT} & \textbf{Extracted Fine-Tuning} & \textbf{SFT+ICT}\\ \midrule
%    Layer 12 & 0.00 & 7.31 & 10.68 & 0.18 \\
%    Layer 16 & 0.12 & 11.14 & - & - \\
%    Layer 20 & 1.83 & 15.83 & 17.00 & 3.99 \\
%    Layer 24 & 1.76 & 20.64 & - & - \\
%    Layer 28 & 3.50 & 25.42 & 28.02 & 6.40 \\
%    Layer 32 & 8.00 & 30.29 & - & - \\
%    Layer 36 & 22.17 & 33.88 & 39.46 & 29.24 \\
%    Layer 40 & 40.6 & 37.96 & - & - \\
%    \bottomrule
%  \end{tabular}}
%  % \vspace{-3mm}
%  \caption{
%   \small{The results on TriviaQA benchmark. Reported numbers are case-sensitive exact match accuracy.}
%   % \vspace{-4mm}
%  }
%  \label{tab:triviaqa}
% \end{table}

\vspace{-4mm}
\subsection{How does SoFT work for other domains?}
We further evaluated Sorted LLaMA in a different domain from the instruction following, selecting the TriviaQA \cite{JoshiTriviaQA2017} 
% and GSM8K \cite{Cobbe2021TrainingVT}
benchmark to assess the sub-models performance in open-domain closed-book questions answering. %Table \ref{tab:gsm8k_results}

Figure \ref{fig:triviaqa} shows the performance of SoFT and three baselines (SFT, Extracted Fine-Tuning and SFT+ICT) in different checkpoints through the training procedure on the TriviaQA benchmark. SoFT sub-models show significant superior performance compared to SFT and SFT+ICT counterparts in all sub-models. Similar to PandaLM, the gap between SoFT and SFT full-model performance is small in TriviaQA, which can underscore the SoFT capability in maintaining full-model performance compared to SFT. We also did Extracted Fine-Tuning on intermediate sub-models for 2 Epochs and results demonstrate close performance of SoFT intermediate layers to Extracted Fine-Tuning counterparts.

\subsection{How can SoFT accelerate text generation?}
\paragraph{Improving Speculative Sampling}
Speculative Decoding (SD) is a technique introduced by \cite{chen2023accelerating} to increase the speed of text decoding in large models. The method utilizes a large target and smaller draft models to generate tokens faster. We can verify the generated tokens by the large model in parallel. We used the same paradigm for Sorted LLaMA as we used earlier sub-models as draft and the full-size model as the target model. As the parameters have been shared between the large and draft models in this setup, we can avoid any extra memory overhead, unlike the standard Speculative Sampling. Table \ref{tab:speculative} reports the inference results of using speculative decoding on Alpaca and TriviaQA benchmarks in SoFT by using three different sub-models as drafts (Layer 12, 16, and 20). As shown combining Speculative decoding and Sorted LLaMA can speed up the token generation up to 1.16$\times$ compared to normal auto-regressive decoding in PandaLM with negligible performance drop. Due to the short average length of answers in TriviaQA, speculative decoding does not result in speed up as the draft generation process does not find any opportunity to accelerate inference.

\begin{figure*}[!htb]
\centering
% \resizebox{\textwidth}{!}{
% \begin{subfigure}[b]{1.1\textwidth}
%    \includegraphics[width=1\linewidth]{Figures/sftAll_sftLast_all.pdf}
%    \caption{LLaMA sub-models vs LLaMA Last Layer}
%     \label{subfig:sft_sft}
% \end{subfigure}

% \begin{subfigure}[b]{1.1\textwidth}
%    \includegraphics[width=1\linewidth]{Figures/sortedAll_sortedLast_all.pdf}
%    \caption{SortedLLaMA sub-models - SortedLLaMA Last Layer}
%    \label{subfig:sorted_sorted}
% \end{subfigure}

\begin{subfigure}[b]{1.05\textwidth}
   \includegraphics[width=1\linewidth]{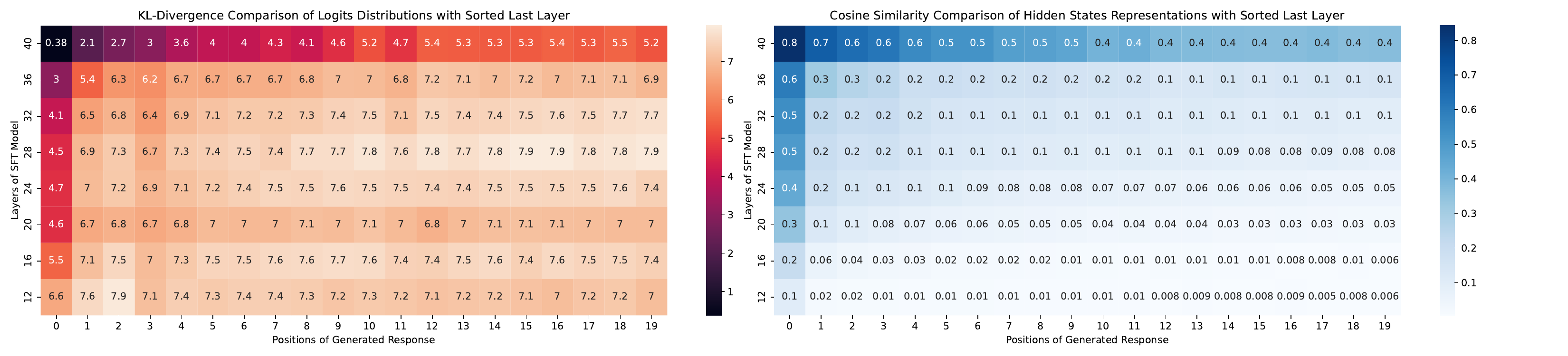}
   \caption{SortedLLaMA Last Layer - LLaMA sub-models}
   \label{subfig:sft_all_sorted_last}
\end{subfigure}

\begin{subfigure}[b]{1.05\textwidth}
   \includegraphics[width=1\linewidth]{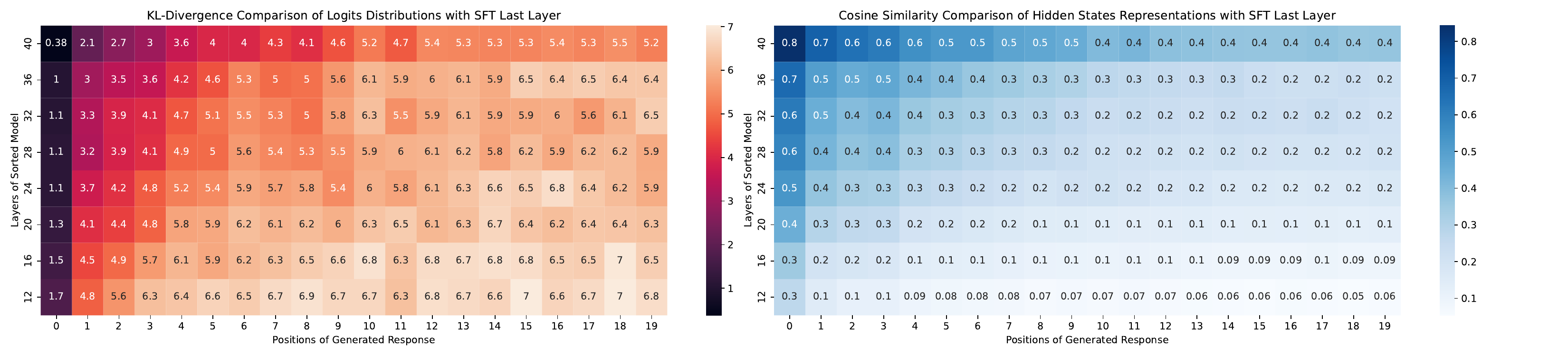}
   \caption{SortedLLaMA sub-models - LLaMA Last Layer}
   \label{subfig:sorted_all_sft_last}
\end{subfigure}
\caption{\small{An inter-model comparison of sub-models based on output logits and hidden state cosine similarity. The numbers are average of all 170 samples in the PandaLM validation set. The similarity is stronger if the cell is darker.}}
\label{fig:statistical_analysis}
\end{figure*}

\paragraph{Instance-Aware Dynamic Inference}
We also dynamically utilize SoFT sub-models to increase the speed of text generation during inference. Based on the confidence of the sub-model's predicted tokens, we decide which sub-model should generate which token. Given each token, the sub-models would process the token in size order (first smallest sub-model 12, then 16, and so on). Wherever in this procedure, the confidence of the predicted token by a sub-model reach a pre-defined threshold, the predicted token would be chosen as the next token and exit the model. We also implemented an adaptive caching mechanism in order to utilize KV caching in this non-trivial scenario where each token can exit from a different layer. Table \ref{tab:speculative} shows that Instance-Aware Dynamic Inference can speed up the normal auto-regressive approach in all benchmarks up to 1.34$\times$ in PandaLM and 1.12$\times$ in TriviaQA. Furthermore dynamic inference can result in better performance in PandaLM and TriviaQA compared to speculative decoding.
% We believe this happens since there is no target model to verify the generated tokens from intermediate layers, this approach can suffer from performance drop especially in the benchmarks like GSM8K where the evaluation metric is exact match accuracy after generating reasoning steps.

% We also need to mention that we did not use the caching mechanism (Key-Value caching) for dynamic inference since the number of layers each token traverse to get generated is different. We also did the auto-regressive experiment in Table \ref{tab:dynamice_inference} without caching to make the comparison fair. NOT SURE IF IT's a good idea to write this because it's an unverified claim on our side. We never did experimented ourselves. This is quite someone else claim. Why we need to mention the cache in the first place?

% \begin{table}[t]
%  \setlength\extrarowheight{2pt}
%  \centering
%  \scalebox{0.85}{\begin{tabular}{l|cc}
  
%   % \cline{1-3}
%   \toprule
%    \multicolumn{3}{c}{\textbf{Auto-regressive Decoding}} \\
%    \textbf{Model} & \textbf{Time per Token (ms)} & \textbf{Accuracy} \\ 
%    Layer 40 (full) & 140.76 & 75.28 \\
%    \midrule
%    \multicolumn{3}{c}{\textbf{Instance-Aware Dynamic Inference}} \\
%    \textbf{Model} & \textbf{Time per Token (ms)} & \textbf{Accuracy}\\ %\midrule 
%    Layer 12:40 & 69.27 (2.03\times) & 74.98\\
%    \bottomrule
%  \end{tabular}}
%  % \vspace{-3mm}
%  \caption{
%   \small{\rtwo{Speed-up in inference time on GSM8K benchmark achieved by Instance-Aware Dynamic Inference.} 
%   }
%   % \vspace{-4mm}
%  }
%  \label{tab:dynamice_inference}
% \end{table}
\vspace{-3mm}
\subsection{Analysis}
\subsubsection{A comparison between the learned probability distribution of SoFT versus SFT}
Sorted tuning aims to make sub-models performance similar to the full model. To explore the efficacy of the SoFT in closing the gap between sub-models and the full model in instruction following task, we measure the similarity between probability distributions of each token in each sub-model versus the full model using the Kullback–Leibler (KL) divergence. Figure \ref{fig:statistical_analysis} (Left) compares the probability distribution of Sorted LLaMA and SFT sub-models at different output positions. 
%We used Kullback–Leibler(KL)-divergence as a metric to measure the similarity between two probability distributions. 
 %, at least in the initial positions of the generated text. 

Figure \ref{subfig:sft_all_sorted_last} (Left) compares different SFT layers and the last Sorted LLaMA layer. The figure shows that only SFT's full-size output distribution is close to the sorted full-size model, while the other layers' distribution diverges faster in the initial steps compared to the SoFT. This is expected as the language model head is unfamiliar with the learned representation of the middle layers in SFT. In the next section, we compared the learned representations of different sub-models to understand SoFT's impact better.

Figure \ref{subfig:sorted_all_sft_last} (Left) compares the output distribution of all sorted layers to the last SFT layer. Compared to Figure \ref{subfig:sft_all_sorted_last} (Left), Figure \ref{subfig:sorted_all_sft_last} (Left) Sorted LLaMA can preserve the output distribution close to the SFT full-size model even in lower layers for initial output tokens.

\begin{figure*}
\centering
% \resizebox{\textwidth}{!}{
\begin{subfigure}[b]{1.05\textwidth}
   \includegraphics[width=1\linewidth]{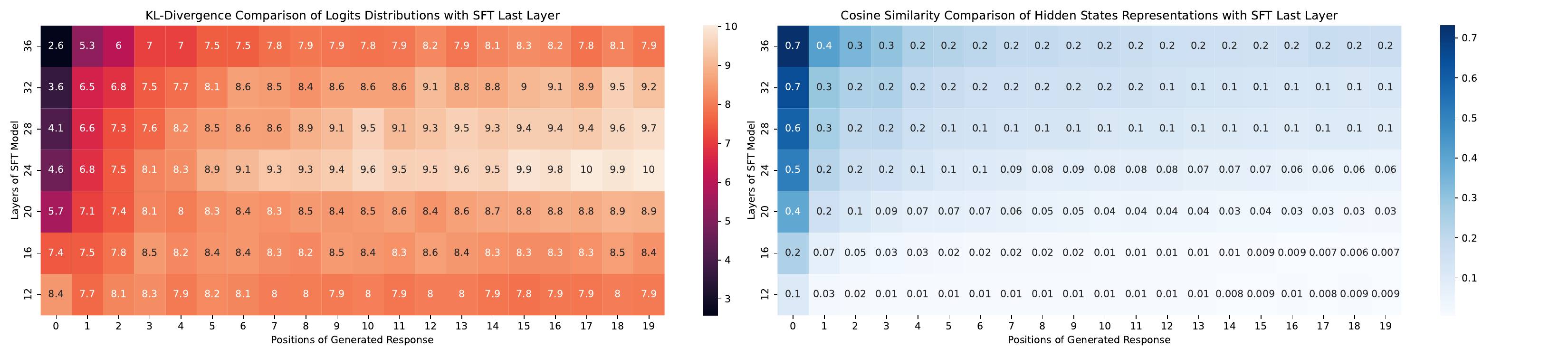}
   \caption{LLaMA sub-models vs LLaMA Last Layer}
    \label{subfig:sft_sft}
\end{subfigure}

\begin{subfigure}[b]{1.05\textwidth}
   \includegraphics[width=1\linewidth]{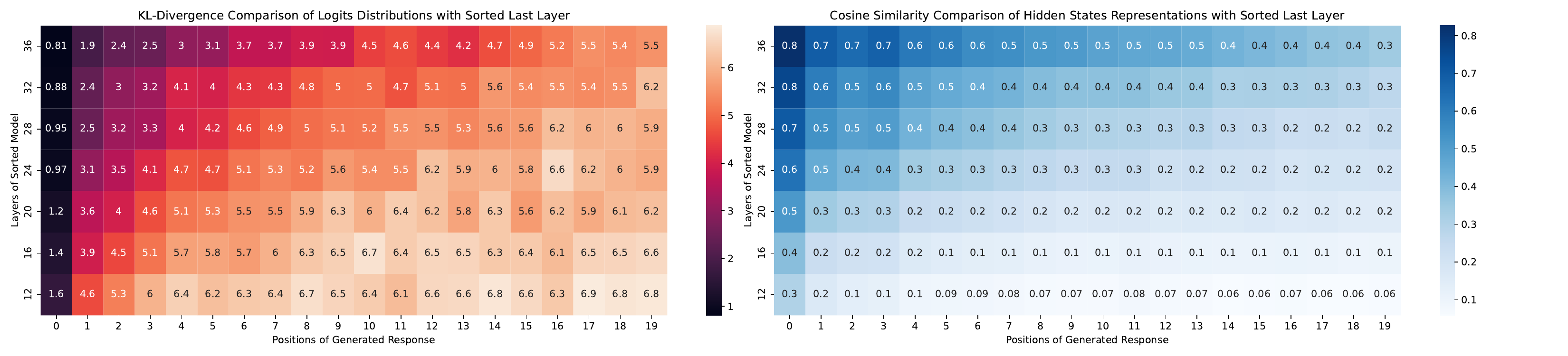}
   \caption{SortedLLaMA sub-models - SortedLLaMA Last Layer}
   \label{subfig:sorted_sorted}
\end{subfigure}

% \begin{subfigure}[b]{1.1\textwidth}
%    \includegraphics[width=1\linewidth]{Figures/sftAll_sortedLast_all.pdf}
%    \caption{SortedLLaMA Last Layer - LLaMA sub-models}
%    \label{subfig:sft_all_sorted_last}
% \end{subfigure}

% \begin{subfigure}[b]{1.1\textwidth}
%    \includegraphics[width=1\linewidth]{Figures/sortedAll_sftLast_all.pdf}
%    \caption{SortedLLaMA sub-models - LLaMA Last Layer}
%    \label{subfig:sorted_all_sft_last}
% \end{subfigure}
\caption{\small{An intra-model comparison of sub-models based on output logits and hidden state cosine similarity. The similarity is stronger if the cell is darker.}}
\label{fig:statistical_analysis_appndx}
\end{figure*}

The comparison between the last layer and the layers 12 to 36 in the SFT model is shown in Figure \ref{subfig:sft_sft} (Left). It is clear from this figure that the output distribution diverges quickly compared to the last layer after generating a few initial tokens, even in higher layers like 36 and 32. It is important to note that this evaluation was generated without adjusting the classifier head.

Finally, Figure \ref{subfig:sorted_sorted} (Left) demonstrates that in Sorted LLaMA, the likelihood distribution of the produced outcome becomes increasingly more similar to the full-size model as we get closer to the last layer.

\subsubsection{A comparison between the learned representation of SoFT versus SFT}

% \todo{how we conducted the experiment? we have to report that this is the average}
%In this part, we conducted another experiment to ensure the misalignment between the language model head and the intermediate layer's learned representation in the SFT setup will not impact our understanding. 
During regular fine-tuning, no connection between the language model head and sub-models can intensify the divergence of probability distributions in Figure \ref{fig:statistical_analysis} (Left). To overcome this, we conducted another experiment to compare the hidden state representation in the last and middle layers just before passing the hidden states to the language model head.
Figure \ref{fig:statistical_analysis} (Right) compares the learned hidden state representation of SFT and Sorted LLaMA sub-models at various positions in the output. This will make the analysis independent of the language model head. We used cosine similarity to measure the difference between the two representations. As shown using heatmaps, the cosine similarities are highly correlated to the KL-Divergence comparison explained in the previous section. 

Figure \ref{subfig:sft_all_sorted_last} (Right) compares all SFT sub-models with the Sorted last layer regarding hidden representation similarity. Again, similar to probability distribution analysis, the similarity between the SFT sub-model and Sorted last layer tends to fade immediately after generating the first few tokens, while Figure \ref{subfig:sorted_all_sft_last} demonstrates the capability of Sorted LLaMA sub-models in preserving the learned representations closely similar to the SFT last layer hidden states.

Figure \ref{subfig:sft_sft} (Right) depicts the heatmap of hidden states cosine similarity among different SFT sub-models compared to the SFT last layer. Similar to its left plot, the similarity quickly diminishes after a few tokens, and this fade is more considerable in earlier layers. 

On the other hand, Figure \ref{subfig:sorted_sorted} (Right) shows that the representations of Sorted sub-models stay similar to the Sorted last layer even after generating multiple initial tokens.

\section{Conclusion}

 %This work aims at training a many-in-one LLM using the SortedNet approach. In this regards, we deployed the SortedNet algorithm to train many-in-one LLaMA 13B model. %Our results showed the effectiveness of our solution. 
This work presents sorted LLaMA, a many-in-one language model for dynamic inference obtained using Sorted Fine-Tuning (SoFT) instead of Standard Fine-tuning. Sorted LLaMA unlocks the potential capability of intermediate layers, offering dynamic adaptation without pre-training or additional costs related to model compression. It presents a promising avenue for optimizing generative language models in NLP. Our approach makes the deployment of these models more efficient. As all sub-models remain integral components of the original model, the burden of storage requirements and transition costs between different computational demands is minimized, making the management of multiple models during inference a practical reality.

Our systematic evaluation of instruction following and questions answering benchmarks challenged conventional wisdom by empowering middle layers to produce high-quality results. This, in turn, enables dynamic inference of LLMs with a highly efficient tuning method (SoFT), ultimately optimizing the usage of LLMs. Our encouraging results show the promising capability of SortedNet \cite{valipour2023sortednet} to train multiple language models with different sizes at once.

\section{Limitations}

Despite showing the effectiveness of the SortedNet approach for large language models, further research is necessary to better understand the scope of its applicability in LLMs. For example, applying this method during pre-training, sorting other model dimensions such as attention heads and hidden dimensions, and investigating the impact of choosing a specific architecture could offer potential avenues for future research. Our study might be slightly biased to automated evaluation, requiring further investigation through human evaluation.

\section*{Acknowledgements}

We thank Mindspore, which is a new deep learning
computing framework, for partial support of this
work. 

% \newpage

\bibliography{custom}

\clearpage
\newpage

\appendix

\section{Appendix}
\label{sec:appendix}

\subsection{Experimental Setup}
\label{appx:experimental_setup}
% \vspace{-0.223cm}
We used the pre-trained LLaMA2 13b weights, publicly available on Hugging Face, as our starting point. For SFT+ICT (Early-Exit) setup, we froze the parameters of the transformer blocks and only further trained the weights of the language model head classifier for one additional epoch. We used a batch size of 32 and gradient accumulation of 8. The learning scheduler was cosine annealing. The learning rate was set to 2e-5 and seed to 42. We trained the models on 8 V100 32GB GPUs. The same GPUs were used during inference time. The training maximum input sequence length was 2024, with a maximum of 50 (TriviaQA) and 256 (PandaLM) generated tokens during inference. Additionally, we used greedy search as the decoding strategy in all of our experiments.
We also extended the huggingface assisted decoding code to implement Speculative Decoding and Instance-Aware Adaptive Inference. In Speculative Decoding, we used adaptive K window-size (the same as huggingface) starting with K=4. In Instance-Aware Dynamic Inference, we set the confidence thresholds of intermediate layers as follow: Layer 12 = 0.95, Layer 16 = 0.95, Layer 20 = 0.9, Layer 24 = 0.9, Layer 28 = 0.8, Layer 32 = 0.8 and Layer 36 = 0.7. 

\subsection{Computational Overhead of SoFT}
\label{appx:comp_overhead}

\begin{table}[ht]
 \setlength\extrarowheight{2pt}
 \centering
 \scalebox{0.62}{\begin{tabular}{l|cc}
  
  % \cline{1-3}
  \toprule
   \textbf{Method} & \textbf{Avg Time per Epoch (s)} & \textbf{Avg Memory Usage per Epoch (MB)} \\ 
   SFT & 25,765.95 & 99,168 \\
   SoFT & 25,269.87 (0.98$\times$) & 125,682\\
   \bottomrule
 \end{tabular}}
 % \vspace{-3mm}
 \caption{
  \small{Training Time and Memory Usage comparison of SoFT and SFT on Alpaca dataset. 
  }
  % \vspace{-4mm}
 }
 \label{tab:computation_overhead}
\end{table}

Given the nested pattern of sub-models and the fact that we share the language model head across sub-models, we do not expect to see any computation overhead for SoFT versus SFT. To validate this claim, we compared SoFT and SFT regarding training time and memory usage in our experiment on the Alpaca dataset. Table \ref{tab:computation_overhead} shows the result for two main experiments of SoFT and SFT. As expected, training with SoFT leads to equal training time compared to SFT. During training, SoFT has about 25\% memory overhead in PyTorch compared to SFT, which only provides a single full model at the end.

\subsubsection{Case Specific Analysis}

Table \ref{tab:examples} shows a sample of instructions from the PandaLM benchmark and the generated responses by SFT+ICT (Early-Exit) and Sorted LLaMA sub-models. 
% In the first example,
Sorted LLaMA performs better in preserving and transferring the last layer performance to earlier sub-models based on the information made visible by black (related to the query) and red (hallucinations, irrelevant, etc.) colors.

Sorted sub-models generate almost correct answers from the 20 layers sub-model, while the first meaningful result from SFT+ICT sub-models appears in layer 28. Other samples generated by SoFT and Early-Exit can be found in \ref{appx:analysis}.

\subsection{Additional Experiments}
\label{appx:additional_experiments}
Table \ref{tab:pandalm_results} shows the detailed results of the Sorted LLaMA and SFT performance on the PandaLM benchmark in different setup in equal training time (2 Epochs for both SFT and SoFT). As we can see, sorted sub-models outperform their SFT counterparts (and even higher sub-models) , while in SFT+ICT (Early-Exit), as we go higher in sub-models (e.g. layer 36), we can see a noticeable improvement in the performance compared to the SFT. This can demonstrate the importance of tuning the language model classifier in improving text generation capability in the latest layers in the standard fine-tuning format. 
% \rgen{SoFT+ICT \st{Early Exit Sorted LLaMA}} shows very similar performance to \rgen{SoFT\st{no Early Exit Sorted LLaMA}} as expected (Figure \ \ref{fig:earlyexitearlyexit}); since adjusting language model classifier does not help sorted sub-models significantly when the classifier has already learned to work with the output of these sub-models during training (Section \ref{sec:methodology}).

Table \ref{tab:pandalm_results_sft_SoFT_4epochs} shows the SoFT and SFT comparison in a different training time setup in which SoFT has access to doubled training time (4 Epochs). Results show that Sorted LLaMA can outperform standard fine-tuned LLaMA further by continuing the SoFT process. The improvement in Sorted LLaMA sub-models performance can be observed specifically in intermediate layers.
% In Table \ref{tab:pandalm_results_sft_SoFT_4epochs}, the performance of zero-shot sorted and SFT with early exit further fine-tuning is shown.  

% Moreover, Table \ref{tab:pandalm_results} and Figure \ref{fig:earlyexitearlyexit} show the results for Sorted LLaMA and SFT sub-models, both in the Early Exit paradigm. Early Exit Sorted LLaMA shows very similar performance to no Early Exit Sorted LLaMA (Figure \ref{fig:earlyexitearlyexit}) as expected; since adjusting language model classifier does not help sorted sub-models significantly when the classifier has already learned to work with the output of these sub-models during training (Section \ref{sec:methodology}).  

\subsection{Analysis}
\label{appx:analysis}

% \begin{figure}[hbt!]
% \hspace*{-1cm}
% \centering
% \resizebox{\columnwidth}{!}{
%         \input{EarlyExitSortedEarlyExitSFTHeatmap.tex}
% }
% \caption{\rgen{SoFT+ICT \st{Early Exit SoFT}} vs. \rgen{SFT+ICT \st{Early Exit SFT}}. To have a fair comparison with \rgen{SFT+ICT \st{Early-Exit}}, our SoFT method has dedicated prediction heads per sub-model. We trained both SFT and SoFT for two epochs in this experiment.}
% \label{fig:earlyexitearlyexit}
% \end{figure}

% \begin{figure}[hbt!]
% \hspace*{-1cm}
% \centering
% % \resizebox{0.5\columnwidth}{!}{
% %         \input{zeroshot4epoSortedCapacitySFT2epoHeatmap}
% % }
% \resizebox{0.1\columnwidth}{!}{
%     \begin{subfigure}[t]{0.2\columnwidth}
%         \input{zeroshot4epoSortedCapacitySFT2epoHeatmap}
%         \label{subfig:zeroshotcapacity4vs2}
%     \end{subfigure}
%     }%
%     % ~
%     \resizebox{0.5\columnwidth}{!}{
%     \begin{subfigure}[t]{0.2\columnwidth}
%             \input{zeroshot4epoSortedCapacitySFT2epoHeatmap}
%             \label{subfig:zeroshotcapacity2vs2}
%     \end{subfigure}
%     }
% \caption{Zero Shot SoFT vs. Capacity SFT.}
% \label{fig:zeroshotcapacity}
% \end{figure}

Table \ref{tab:appx_examples} and \ref{tab:appx_examples_2} show some samples generated by sub-models of LLaMA (SFT+ICT) and SoFT on PandaLM evaluation set. In the first query of Table \ref{tab:appx_examples}, LLaMA sub-models until layer 36 struggle to generate relevant responses about books in the Crime and Mystery genre. Sorted LLaMA sub-models, however, start to address the related novels from layer 24. The second query in the table is a simpler instruction, which is a multi-label classification problem. Again Sorted LLaMA sub-models start to generate the correct label in much earlier layers (layer 20) compared to the LLaMA sub-models (layer 24). Table \ref{tab:appx_examples_2} first example shows the performance gap of the LLaMA and Sorted LLaMA intermediate sub-models even in a more severe case. To write a review about a restaurant with certain aspects, LLaMA sub-models before layer 32 hallucinate or generate gibberish, while Sorted LLaMA starts to generate a complete review addressing key points mentioned in the instruction even in the first sub-model (layer 16). In the second example, the same pattern occurs where SoFT sub-models can generate meaningful response starting from layer 16 while LLaMA first reasonable text happens at layer 36.

Table \ref{tab:appx_example_triviaqa} shows an example of SFT and SoFT performance on TriviaQA benchmark. While LLaMA struggles to generate single answer token even in the sub-models close to the last layer, SoFT could transfer the question answering capability of LLM until sub-layer 20 and still generate the correct final answer.

After all, Sorted LLaMA sub-models demonstrate the ability to generate more comprehensive (Example 1 of Table \ref{tab:appx_examples} and Table \ref{tab:appx_examples_2}) and informative (Table \ref{tab:appx_examples} example 2) answers in earlier layers compared to LLaMA. Based on our observation, LLaMA sub-models mostly tend to generate irrelevant or even gibberish in earlier blocks (layers 12 to 24), while the generated texts by Sorted LLaMA exhibit sufficient learned information to answer the input instruction despite having much fewer parameters.  

\begin{table*}[!ht]
    \centering
    \resizebox{0.9\textwidth}{!}{
    \begin{tabular}{l|l|l|l|l|l|l|l|l}
    \toprule
        Sorted LLaMA/LLaMA & 12 (4.1B) & 16 (5.4B) & 20 (6.6B) & 24 (7.9B) & 28 (9.2B) & 32 (10.4B) & 36 (11.7B) & 40 (13B) \\ \midrule
        \multicolumn{9}{c}{SoFT vs. SFT} \\ \midrule
        12 (4.1B) & 71.0/99.0/0.0 & 97.5/72.5/0.0 & 129.0/41.0/0.0 & 131.0/39.0/0.0 & 121.5/48.5/0.0 & 106.5/63.5/0.0 & 45.0/125.0/0.0 & 17.0/152.5/0.5 \\
        16 (5.4B) & 81.0/89.0/0.0 & 101.5/68.5/0.0 & 128.5/40.5/1.0 & 131.5/38.0/0.5 & 124.0/44.5/1.5 & 114.0/54.0/2.0 & 52.0/114.0/4.0 & 18.0/146.0/6.0 \\
        20 (6.6B) & 111.5/58.5/0.0 & 132.0/38.0/0.0 & 144.5/23.5/2.0 & 147.5/20.5/2.0 & 141.5/24.0/4.5 & 132.5/30.5/7.0 & 73.5/85.5/11.0 & 32.5/114.0/23.5 \\
        24 (7.9B) & 124.5/45.5/0.0 & 136.5/33.5/0.0 & 150.0/18.0/2.0 & 154.5/13.5/2.0 & 148.0/18.5/3.5 & 144.5/19.0/6.5 & 98.0/62.0/10.0 & 44.5/90.0/35.5 \\
        28 (9.2B) & 125.5/44.5/0.0 & 145.0/25.0/0.0 & 153.0/15.0/2.0 & 153.5/14.5/2.0 & 148.0/16.5/5.5 & 143.5/20.5/6.0 & 96.5/59.5/14.0 & 45.0/89.0/36.0 \\
        32 (10.4B) & 141.5/28.5/0.0 & 152.0/18.0/0.0 & 159.0/9.0/2.0 & 160.0/8.0/2.0 & 152.0/12.5/5.5 & 150.5/13.5/6.0 & 108.5/45.0/16.5 & 55.5/75.0/39.5 \\
        36 (11.7B) & 141.0/28.5/0.5 & 152.5/17.0/0.5 & 159.0/8.5/2.5 & 161.5/6.5/2.0 & 150.0/14.5/5.5 & 148.5/15.5/6.0 & 112.0/42.5/15.5 & 53.0/66.0/51.0 \\
        40 (13B) & 143.5/26.5/0.0 & 156.0/14.0/0.0 & 160.5/7.5/2.0 & 161.0/7.0/2.0 & 150.0/14.0/6.0 & 150.0/13.5/6.5 & 115.5/39.0/15.5 & 52.5/62.5/55.0 \\ \midrule
        \multicolumn{9}{c}{SoFT vs. SFT+ICT(Early-Exit)} \\ \midrule
        12 (4.1B) & 75.0/95.0/0.0 & 108.5/61.5/0.0 & 128.5/41.5/0.0 & 122.5/47.5/0.0 & 116.5/53.5/0.0 & 91.0/79.0/0.0 & 37.5/131.5/1.0 &  17.0/152.5/0.5 \\
        16 (5.4B) & 86.5/82.5/1.0 & 113.0/57.0/0.0 & 127.0/41.0/2.0 & 122.0/47.0/1.0 & 117.5/50.5/2.0 & 94.5/72.0/3.5 & 36.0/129.0/5.0 & 18.0/146.0/6.0 \\
        20 (6.6B) & 111.5/57.5/1.0 & 137.0/33.0/0.0 & 143.5/24.0/2.5 & 143.0/23.0/4.0 & 137.0/27.0/6.0 & 122.0/38.0/10.0 & 60.0/94.5/15.5 & 32.5/114.0/23.5 \\
        24 (7.9B) & 126.5/42.5/1.0 & 144.0/26.0/0.0 & 149.0/19.5/1.5 & 151.0/15.5/3.5 & 143.0/21.5/5.5 & 133.5/28.0/8.5 & 76.5/72.5/21.0 & 44.5/90.0/35.5 \\
        28 (9.2B) & 130.0/39.0/1.0 & 147.0/23.0/0.0 & 153.5/15.5/1.0 & 150.0/16.0/4.0 & 143.5/18.5/8.0 & 131.0/29.0/10.0 & 79.0/66.0/25.0 & 45.0/89.0/36.0 \\
        32 (10.4B) & 141.5/27.5/1.0 & 155.5/14.5/0.0 & 161.0/8.0/1.0 & 157.0/8.5/4.5 & 151.0/11.0/8.0 & 143.5/15.0/11.5 & 89.5/49.5/31.0 & 55.5/75.0/39.5 \\
        36 (11.7B) & 143.0/25.5/1.5 & 156.5/13.0/0.5 & 160.0/8.5/1.5 & 157.0/8.5/4.5 & 148.0/14.0/8.0 & 142.5/16.5/11.0 & 92.5/46.5/31.0 & 53.0/66.0/51.0 \\
        40 (13B) & 146.0/23.0/1.0 & 157.0/13.0/0.0 & 160.5/7.5/2.0 & 157.5/9.0/3.5 & 149.0/14.0/7.0 & 143.5/16.0/10.5 & 97.5/43.5/29.0 & 52.5/62.5/55.0 \\
        \bottomrule
    \end{tabular}}
    \caption{\small{Pair-wise comparison for different layers (sub-models) in Standard Fine-Tuning and SoFT at equal training cost (2 Epochs). Each cell consists of three values: Wins, Losses, Ties. Wins demonstrate the number of times that the generated text of the sub-model in row (sorted) is preferred to the sub-model in column (Fine-Tuned) and Losses is the opposite. Numbers are average of two separate experiments with different order of inputs to evaluator in order to neutralize the order bias.}}
    \label{tab:pandalm_results}
\end{table*}

\begin{table*}[!ht]
    \centering
    \resizebox{0.9\textwidth}{!}{
    \begin{tabular}{l|l|l|l|l|l|l|l|l}
    \toprule
        Sorted LLaMA/LLaMA & 12 (4.1B) & 16 (5.4B) & 20 (6.6B) & 24 (7.9B) & 28 (9.2B) & 32 (10.4B) & 36 (11.7B) & 40 (13B) \\ \midrule
        \multicolumn{9}{c}{SoFT vs. SFT} \\ \midrule
        12 (4.1B) & 88.5/81.5/0.0 & 108.0/62.0/0.0 & 134.5/35.5/0.0 & 135.0/35.0/0.0 & 129.0/41.0/0.0 & 120.0/49.0/1.0 & 57.0/109.5/3.5 & 23.5/144.0/2.5 \\
        16 (5.4B) & 106.5/63.0/0.5 & 120.0/50.0/0.0 & 140.0/29.0/1.0 & 144.5/24.5/1.0 & 142.0/26.5/1.5 & 136.0/32.0/2.0 & 70.0/95.0/5.0 & 34.5/124.5/11.0 \\
        20 (6.6B) & 127.0/43.0/0.0 & 138.5/31.5/0.0 & 151.5/16.5/2.0 & 152.0/17.0/1.0 & 143.5/23.5/3.0 & 144.0/21.5/4.5 & 94.5/67.5/8.0 & 47.0/99.5/23.5 \\
        24 (7.9B) & 138.5/31.5/0.0 & 149.5/20.5/0.0 & 159.0/9.0/2.0 & 158.0/10.5/1.5 & 151.5/13.5/5.0 & 149.0/15.5/5.5 & 107.0/49.5/13.5 & 53.0/81.0/36.0 \\
        28 (9.2B) & 137.0/33.0/0.0 & 149.0/21.0/0.0 & 158.0/10.0/2.0 & 159.5/8.5/2.0 & 150.0/15.0/5.0 & 149.5/15.0/5.5 & 107.0/47.5/15.5 & 50.5/78.0/41.5 \\
        32 (10.4B) & 146.0/24.0/0.0 & 157.0/13.0/0.0 & 163.0/5.0/2.0 & 163.0/5.0/2.0 & 154.5/10.5/5.0 & 151.5/12.5/6.0 & 117.5/37.5/15.0 & 63.5/62.0/44.5 \\
        36 (11.7B) & 149.5/20.5/0.0 & 160.0/10.0/0.0 & 164.0/4.0/2.0 & 162.5/5.5/2.0 & 157.5/7.5/5.0 & 154.0/10.0/6.0 & 119.5/34.5/16.0 & 62.5/60.0/47.5 \\
        40 (13B) & 153.5/16.5/0.0 & 163.0/7.0/0.0 & 165.5/3.0/1.5 & 163.5/4.5/2.0 & 157.0/8.0/5.0 & 156.0/8.5/5.5 & 121.0/33.5/15.5 & 67.5/52.0/50.5 \\ \midrule
        \multicolumn{9}{c}{SoFT vs. SFT+ICT(Early-Exit)} \\ \midrule
        12 (4.1B) & 91.5/77.5/1.0 & 123.5/46.5/0.0 & 138.5/31.5/0.0 & 134.0/36.0/0.0 & 130.5/39.0/0.5 & 107.5/59.0/3.5 & 46.0/120.5/3.5 & 23.5/144.0/2.5 \\
        16 (5.4B) & 106.5/63.5/0.0 & 128.5/41.0/0.5 & 145.0/24.0/1.0 & 144.5/25.0/0.5 & 139.0/29.5/1.5 & 122.5/43.0/4.5 & 55.5/106.5/8.0 & 34.5/124.5/11.0 \\
        20 (6.6B) & 128.0/40.5/1.5 & 142.0/27.5/0.5 & 152.5/16.0/1.5 & 148.0/19.0/3.0 & 142.0/22.5/5.5 & 131.0/31.0/8.0 & 74.0/79.0/17.0 & 47.0/99.5/23.5 \\
        24 (7.9B) & 140.5/28.5/1.0 & 155.0/15.0/0.0 & 159.0/10.0/1.0 & 156.0/10.5/3.5 & 149.5/12.5/8.0 & 141.5/19.0/9.5 & 90.0/54.5/25.5 &  53.0/81.0/36.0 \\
        28 (9.2B) & 142.0/27.0/1.0 & 155.5/14.5/0.0 & 161.5/7.5/1.0 & 155.5/10.5/4.0 & 148.5/13.0/8.5 & 141.0/19.0/10.0 & 91.5/53.5/25.0 & 50.5/78.0/41.5 \\
        32 (10.4B) & 150.0/19.0/1.0 & 161.0/9.0/0.0 & 161.5/7.5/1.0 & 160.0/6.0/4.0 & 153.0/8.5/8.5 & 145.0/14.0/11.0 & 99.0/41.0/30.0 & 63.5/62.0/44.5 \\
        36 (11.7B) & 151.0/18.0/1.0 & 163.5/6.5/0.0 & 165.0/3.5/1.5 & 161.0/5.0/4.0 & 155.0/7.5/7.5 & 147.5/12.0/10.5 & 105.5/36.0/28.5 & 62.5/60.0/47.5 \\
        40 (13B) & 156.0/13.0/1.0 & 165.5/4.5/0.0 & 165.0/3.5/1.5 & 161.0/5.0/4.0 & 155.0/7.5/7.5 & 150.5/10.0/9.5 & 106.0/35.0/29.0 & 67.5/52.0/50.5 \\
        \bottomrule
    \end{tabular}}
    \caption{\small{Pair-wise comparison for different layers (sub-models) in Standard Fine-Tuning and SoFT (4 Epochs) after tuning Language Model classifier for each SFT sub-model.}}
    \label{tab:pandalm_results_sft_SoFT_4epochs}
\end{table*}

\begin{table*}[htb!]

\tiny
%    \begin{tabularx}{\textwidth}{|p{0.1\textwidth} | p{0.2\textwidth} | p{0.25\textwidth} | p{0.2\textwidth} | p{0.25\textwidth} |}
% \resizebox{\textwidth}{!}{
\centering
\scalebox{1}{
    \begin{tabularx}{\textwidth}{|l | X | X |}

    \toprule
    % \multicolumn{3}{|c|}{Query:} \\ \hline
    \multicolumn{3}{|l|}{\makecell[l]{\textbf{Query}: Choose the best books from the given genre. Crime \& Mystery}} \\ \midrule
    % \multicolumn{3}{|c|}{Responses:} \\ \hline
    \textbf{Layer} & \textbf{LLaMA Response (SFT + ICT)} & \textbf{Sorted LLaMA Response (SoFT)} \\ \midrule 

        12 & \textcolor{purple}{Therewiguptaranapped byoremandoesleven years ago agomairiensum...} & \color{purple}{Here are some of the best books from the genre: 1. The Fear of the unknown: The Fist of the Darkness2. The Fist of the Deaths of the Deaths of the Deaths of the Deaths of the Deaths of the Deaths ...}  \\ \hline

        16 & \textcolor{purple}{Here list list list!!!!!!!!!!!!!!!\u00e4ndersonally preferences personal ...} & \color{purple}{Here is a list of top 10 books from the genre  Crime \& Mystery that you can read for your next reading book list:  1. The Great mystery of the Mystery of the Mystery of the Mystery of the Mystery of the ...} \\ \hline

        20 & \textcolor{purple}{Here listicle lists some best books genreed intoxicatedly intoxicatedly ...} & Here are some popular and well-known crime and mystery books that you can choose from:  1. The Secret of the Mystery solved by the detective by Jameson Jameson is a classic crime and mystery novel that is widely regarded as one of the best crime and mystery books ever written.  \color{purple}{2. The Secret of the Mystery solved by the detective by Jameson Jameson is a classic crime and mystery novel that is widely regarded as one of the best crime and mystery books ever written.  3. The Secret of the Mystery solved ...} \\ \hline

        24 & \textcolor{purple}{Here list contains some best books belonging belongs genre crime \& mystery genre ordered alphabetically alphabetically ...} & Here are some of the best crime and mystery books:  1. The Hatchery by Stephen King - This book is a classic horror/ crime thriller that follows the story of a group of friends who discover a hidden hatchery of horrors. 2. The Last Thistle by Murphy Murphy - This book is a thriller that follows the investigation of a murder in a small town. The story is filled with suspense and twists that will keep you hooked. \color{purple}{3. The Last Thistle by Murphy Murphy - This book is a thriller that follows the investigation of a murder in a small town. The story is filled with suspense and twists that will keep you hooked. 4. The Last Thistle by Murphy ...}  \\ \hline 

        28 & \textcolor{purple}{crime mystery books best crime mystery books crime mystery books best crime mystery ...} & Here are some of the best crime and mystery books that you might enjoy:  1. The Great Gatsby by F. Scott Fitzgerald - This classic novel is a classic tale of love, betrayal, and the dark side of the American Dream. \color{purple}{2. The Great Gatsby by F. Scott Fitzgerald - This classic novel is a classic tale of love, betrayal, and the dark side of the American Dream. 3. The Great Gatsby ...} \\ \hline 

        32 & \textcolor{black}{Crime mystery genre has many great books to choose from. Here are some recommendations:} \color{purple}{crime mystery books recommended by readers: 1) crime mystery books recommended by readers: Crime mystery genre has many great books to choose from. Here are some recommendations: crime mystery books ...} & Crime \& Mystery is a popular genre, and there are many great books that fall into this category. Here are some of the best crime and mystery books that have stood the test of time and are widely acclaimed by readers and critics alike:  1. The Hound of the Baskervilles by Sir Arthur Conan Doyle: This classic detective novel features the iconic detective Sherlock Holmes and his sidekick, Dr. John Watson. The story follows Holmes as he investigates the mysterious death of a wealthy man and the disappearance of his family.  2. The Silent Patient by Sarah C. Dessen: This psychological thriller follows the story of a woman who is institutionized after shooting her husband. The novel explores themes of trauma, mental illness, and the power of love to heal.  3. The Silent majority by Agatha Christie: This classic mystery novel features the beloved detective Hercule Poirot as he investigates the murder of a wealthy man. The novel is full of twists and turns, and the identity of the murderer is not revealed until the end. \\ \hline

        36 & Crime fiction is a genre that has captivated readers for centuries with intriguvity and suspenseful plots. Here are some of the best crime books that you should definitely read: 1. Sherlock Holmes series by Sir Arthur Conan Doyle: regarded as one of the greatest fictional detectives of all time, Sherlock Holmes is known for his deductive reasoning and ability to solve complex cases. 2. Agatha Christie novels: regarded as the queen of crime fiction, Agatha Christie wrote numerous novels and short stories that are still widely read today. Some of her most famous works include Hercules Poirot series and Miss Marple series. Raymond Chandler novels: regarded as one of the greatest crime writers of all time, Raymond Chandler wrote novels set in Los Angeles featuring detective Philip Marlowe. Some of his most famous works include \"The Big Sleep\" and \"The Long Goodbye\". \color{purple}{4. crime novels by crime novelist Raymond Chandler: regarded as one of the greatest crime writers of all time, Raymond Chandler wrote novels set in Los Angeles featuring detective Philip Marlowe. Some of his most famous} & Based on your preference for crime and mystery books, here are some highly recommended titles:  1. To Kill a Mockingbird by Harper Lee: This classic novel explores themes of racism, prejudice, and injustice in the Deep South during the 1930s.  2. The Silent Patient by Alex Michaelides: A psychological thriller about a woman who shoots her husband and then refuses to speak.  3. The Girl with the Dragon Tattoo by Stieg Larsson: This novel kicks off the Millennium Trilogy, following the investigation of a missing journalist and the mysterious disappearance of a young woman.  4. The Da Vinci Code by Dan Brown: This fast-paced thriller follows the trail of a centuries-old conspiracy involving the Catholic Church, the Knights Templar, and the secrets of the Da Vinci Code.  5. The Girl with the Dragon Tattoo by Stieg Larsson: This novel kicks off the Millennium Trilogy, following the investigation of a missing journalist and the mysterious \\ \hline 

        40 & Here are some of the best books in the Crime and Mystery genre:  1. The Hound of the Baskervilles by Sir Arthur Conan Doyle 2. The Maltese Falcon by Dashiell Hammett 3. The Big Sleep by Raymond Chandler 4. The Adventures of Sherlock Holmes by Sir Arthur Conan Doyle 5. The Thin Man by Dashiell Hammett 6. The Murder of Roger Ackroyd by Agatha Christie 7. The Silence of the Lambs by Thomas Harris 8. The Red House Mystery by A.A. Milne 9. The Secret of the Old Clock by Carolyn Keene 10. The Adventures of the Black Stallion by Walter Farley. & Here are some highly recommended crime and mystery books:  1. To Kill a Mockingbird by Harper Lee 2. The Hound of the Baskervilles by Sir Arthur Conan Doyle 3. The Silence of the Lambs by Thomas Harris 4. The Great Gatsby by F. Scott Fitzgerald 5. The Catcher in the Rye by J.D. Salinger 6. The Da Vinci Code by Dan Brown 7. The Girl with the Dragon Tattoo by Stieg Larsson 8. The Adventures of Sherlock Holmes by Sir Arthur Conan Doyle 9. The Secret of the Old Clock by Carolyn Keene 10. The Help by Kathryn Stockett. \\ \midrule

        % \multicolumn{3}{|c|}{Query:} \\ \hline
        \multicolumn{3}{|l|}{\makecell[l]{\textbf{Query}: Classify the questions in the FAQ into Finding policy details, Change account settings, Filing a claim and viewing status, or Cancelling coverage. \\
Could you deposit money into my account rather than mailing me a physical cheque?}} \\ \midrule
        % \multicolumn{3}{|c|}{Responses:} \\ \hline
        \textbf{Layer} & \textbf{LLaMA Response (SFT + ICT)} & \textbf{Sorted LLaMA Response (SoFT)} \\ \midrule

        12 & \textcolor{purple}{Yes sir/ sorry sorryonymouslyovoyedance Festivalandeiamot...}& \textcolor{purple}{The question is not classified as a request or request for information. It is a question that is addressed to the customer's request.} \\ \hline

        16 & \textcolor{purple}{belongs belongsfrontieralandersonallyi...} & \textcolor{purple}{The given question is a question of financial accounting and it belongs to the Cancelling coverage category.} \\ \hline

        20 & \textcolor{purple}{belongs category \" changeable via email or phone call or online portal or phone contactless payment etceterially...} & Change account settings \\ \hline

        24 & Change account settings & Change account settings \\ \hline

        28 & This question falls category of Change account settings. & Change account settings \\ \hline

        32 & Change account settings & Change account settings \\ \hline
        
        36 & Change account settings & Change account settings \\ \hline

        40 & Change account settings & Change account settings \\ \bottomrule

    \end{tabularx}}
    \caption{\small{Generated responses in different layers of SFT (Early Exit) and Sorted LLaMA for two samples from PandaLM validation set.}}
    \label{tab:appx_examples}
\end{table*}

\begin{table*}[htb!]

\tiny
%    \begin{tabularx}{\textwidth}{|p{0.1\textwidth} | p{0.2\textwidth} | p{0.25\textwidth} | p{0.2\textwidth} | p{0.25\textwidth} |}
% \resizebox{\textwidth}{!}{
\centering
\scalebox{1}{
    \begin{tabularx}{\textwidth}{|l | X | X |}

    \toprule
         % \multicolumn{3}{|c|}{Query:} \\ \hline
        \multicolumn{3}{|l|}{\makecell[l]{\textbf{Query}: Based on the given keywords, write a review of a restaurant. Name: Pizzeria \\ 
good prices, quiet place, tasty, nice staff}} \\ \midrule
        % \multicolumn{3}{|c|}{Responses:} \\ \hline
        \textbf{Layer} & \textbf{LLaMA Response (SFT + ICT)} & \textbf{Sorted LLaMA Response (SoFT)} \\ \midrule

        12 & \textcolor{purple}{Hinweis B\u00e9jar Brothersqueeries...} & \color{purple}{The restaurant is located in a small town, with an impressive list of options for meals and drinks. The atmosphere is comfortable and comfortable, with a small, large-scale room, with a large enough room for comfortable comfortable rooms. The menu is a great way to choose the best option for meals and drinks. The menu is easy to understand and can be easily used to find the best option for meals and drinks. The menu is also available online and can be easily accessed by anyone who is interested in making a purchase. The menu is also made up of a variety of cuisine, with a variety of cuisine. The menu is also made up of a variety of cuisine...} \\ \hline

        16 & \textcolor{purple}{Review: description description Cassandra Smithsoniusque ...} & The Pizzeria is a great place to eat and enjoy a relaxing meal. It is a quiet place, so it is perfect for a quiet meal and a relaxing meal. The food is incredibly delicious, and the staff is friendly \color{purple}{and friendly.} \color{black}{The prices are great, and the atmosphere is comfortable and relaxing. It is a great place to eat and enjoy a relaxing meal.} \\ \hline

        20 & \textcolor{purple}{Review: visited recently visited recently visited recently visited recently...} & I recently visited Pizzeria, a restaurant that is both affordable and a quiet place to enjoy a meal. The prices are definitely good, and the food is definitely tasteful. The staff is nice and friendly, making the experience even more enjoyable. Overall, I highly recommend Pizzeria for anyone looking for a good meal at a reasonable price. \\ \hline

        24 & \textcolor{purple}{Review: visited recently visited recently visited recently visited recently ...} & I recently visited Pizzeria, and I must say, it was a pleasant experience. The prices were reasonable, and the atmosphere was quiet and relaxing. The staff was friendly and helpful, making me feel welcome and comfortable. The pizza was tasty and made me feel satisfied after a long day. Overall, I would recommend Pizzeria to anyone looking for a good pizza, good prices, and a nice atmosphere \\ \hline

        28 & \textcolor{purple}{Revieweddit rating: stars star star star ...} & Pizzeria is a hidden gem in the heart of the city. The prices are reasonable and the atmosphere is a refreshing change from the crowded restaurants. The staff is friendly and welcoming, making the experience even more enjoyable. The pizza is tasty and the crust is crisp and crunchy. The sauce is rich and the cheese is melted perfectly. Overall, Pizzeria is a must-visit for anyone looking for a good pizza and a quiet place to enjoy it. \\ \hline

        32 & Review: visited this place with friends recently after hearing good reviews about their pizza pie. Upon arrival we noticed how quiet it was inside which made it perfect place to relax after hectic weekdays. prices were reasonable considering quality of food served. staff members were friendly and helpful throughout entire duration of visit. overall experience exceed expectations thanks to delicious food served hot freshly baked pizza \color{purple}{pie tasteful ingunctions of Italian cu cu cu cu cu cu cu cu cu cu ...} & I recently visited Pizzeria and I must say, it was a pleasant experience. The prices were reasonable, and the atmosphere was quiet and serene. The staff was friendly and helpful, making me feel welcome. The pizza was delicious, with a crisp crust and a delicious, creamy sauce. Overall, I would definitely recommend Pizzeria to anyone looking for a good meal in a quiet, relaxing setting. \\ \hline

        36 & Review of Pizzeria: upon walking into pizzeria, you are greetvity welcomed by nice staff members who are always ready to assist you with your needs. Their prices are reasonable considering the quality of food they offer. Their pizza is mouthwateringly delicious and always freshly made to order. Additionally, pizzeria is a quiet place perfect for dates or family gatherings. Overall, pizzeria is a great place to visit for delicious food and great service. highly recommended! & I recently visited Pizzeria and had a wonderful experience. The prices were reasonable and the atmosphere was quiet and relaxing. The food was delicious and the staff was friendly and attentive. Overall, I would highly recommend Pizzeria to anyone looking for a tasty meal in a nice, quiet setting. \\ \hline

        40 & Pizzeria is a great place to go for a quiet meal with tasty food and friendly staff. The prices are good, making it a great value for your money. The atmosphere is cozy and the service is prompt. The pizza is cooked to perfection and the toppings are fresh and delicious. Overall, it's a great place to enjoy a meal with friends or family. Highly recommended! & Pizzeria is a great place to enjoy a delicious pizza in a quiet and cozy atmosphere. The prices are reasonable and the staff is friendly and accommodating. The pizzas are cooked to perfection and the toppings are fresh and tasty. The restaurant is perfect for a casual meal with friends or family, and the ambiance is relaxing and inviting. Overall, Pizzeria is a great option for a satisfying pizza experience.\\ \midrule
         \multicolumn{3}{|l|}{\makecell[l]{\textbf{Query}: Define what the underlined word means for kids. \_keep a promise}} \\ \midrule
        % \multicolumn{3}{|c|}{Responses:} \\ \hline
        \textbf{Layer} & \textbf{LLaMA Response (SFT + ICT)} & \textbf{Sorted LLaMA Response (SoFT)} \\ \midrule

        12 & \color{purple}{Thereuh Scientificallyally compromiseable ...} & \color{purple}{The word machery is a word that is used to describe a person or a person's behavior, or a person's behavior in a particular manner. It means that someone has been able to successfully successfully deal with the situation or situation situationally.} \\ \hline

        16 & \color{purple}{means something importantlyrics song songwriter} & The word \color{purple}{Kick A promise} \color{black}{is a common term in the context of a commitment to deliver something or to take action. It means that someone has taken a promise to make a promise to make sure that they are doing something or fulfilling their promise. It is a kind of commitment that is often used in personal and professional settings.} \\ \hline

        20 & \color{purple}{When someone promise means promise means promise means promise means promise means ...} & \color{purple}{Keeping a promise means keeping a promise made to someone else, and not making a promise to someone else or even yourself that you will not do something, but actually doing it. It means being kind, honesty, and kindness towards others. It means being kind to your parents, siblings, friends, and even your peers. It means being kind to yourself too. It means being kind to yourself ...}\\ \hline

        24 & \color{purple}{When someone promise promise means promise promise means promise ...} & Keep a promise means to make sure that you do what you promised to do. It means that you are reliable and reliable people are people who are trusted and respected. When you promise something to someone, it means that you are committed to it and you will do it. \\ \hline

        28 & \color{purple}{keep promise means promise promise means promise promise means promise promise ...} & Keeping a promise means to do what you promised to do, no matter what. It means keeping your word and keeping your word is important. It's important to keep promises because it helps people trust you and trust that you will do what you say you will do.\\ \hline

        32 & \color{purple}{When you promise something means prom promprompr...} & Keeping a promise means that you promise something to someone and you do it. It means that you do not break your word and you do what you said you would do. It is important to keep promises because it shows that you are trustworthy and reliable.\\ \hline

        36 & \color{black}{Keeping a promise means promising something to someone else or yourself, and then doing your best to make sure that promise comes true. It means being honest and keeping your word even when} \color{purple}{it'vity hard or takes effort.} \color{black}{It means being responsible and reliable so people can trust you to do what you say you will do.} & Keeping a promise means doing what you said you would do. It's important to follow through on your promises to others and to be reliable.\\ \hline

        40 &  Keep a promise means to do what you said you would do. It means to be honest and to keep your word. When you keep a promise, you show that you are trustworthy and that people can count on you. & Keeping a promise means doing what you said you would do, even if it's hard or takes a long time. It's important to be honest and reliable, and to follow through on your words. \\ \bottomrule
        
    \end{tabularx}}
    \caption{\small{Generated responses in different layers of SFT (Early Exit) and Sorted LLaMA for a sample from PandaLM validation set.}}
    \label{tab:appx_examples_2}
\end{table*}

\begin{table*}[hbt!]

\scriptsize
%    \begin{tabularx}{\textwidth}{|p{0.1\textwidth} | p{0.2\textwidth} | p{0.25\textwidth} | p{0.2\textwidth} | p{0.25\textwidth} |}
\resizebox{\textwidth}{!}{
    \begin{tabularx}{\textwidth}{|l | X | X |}

    \toprule
    % \multicolumn{3}{|c|}{Query:} \\ \hline
    \multicolumn{3}{|l|}{\makecell[l]{\makecell[l]{\textbf{Query}: Bandar Seri Begawan International airport is in which country?}}} \\ \midrule 
    \textbf{Layer} & \textbf{LLaMA Response (SFT)} & \textbf{Sorted LLaMA Response (SoFT)} \\ \midrule %& \textbf{SFT} & \textbf{Sorted LLaMA} \\ \hline

        12 & \color{purple}{iernohotter than than  ...} & \color{purple}{Malta}\\ \hline 

        16 & \color{purple}{Nederlige territ territorially ...} & \textcolor{purple}{Burma (Myanmar)}\\ \hline 

        20 & \color{purple}{Australia AustraliaAustral Australia Australia ...} & Brunei\\ \hline 

        24 & \color{purple}{Malays Malays Malays Malays ...} & Brunei\\ \hline 
        
        28 & \color{purple}{Malays Malays Malays Malays ...} & Brunei\\ \hline 

        32 & Brunei & Brunei\\ \hline 

        36 & Brunei & Brunei\\ \hline 

        40 & Brunei & Brunei 
        \\ \bottomrule
        
    \end{tabularx}}
    \caption{\small{Generated responses in different layers of SFT and SoFT for a sample from TriviaQA benchmark.}}
    \label{tab:appx_example_triviaqa}
\end{table*}

\begin{table*}[!ht]
    \centering
    \begin{tabular}{l|l|l|l|l}
    \toprule
        Sorted/Fully Fine-tuned & 12 (4.1B) & 20 (6.6B) & 28 (9.2B) & 36 (11.7B) \\ \midrule
        \multicolumn{5}{c}{2 SFT Epochs/2 SoFT Epochs} \\ \midrule
        12 (4.1B) & 80.0/88.5/1.5 & 37.5/132.0/0.5 & 28.0/141.5/0.5 & 20.0/148.5/1.5 \\
        16 (5.4B) & 88.5/77.0/4.5 & 42.0/121.5/6.5 & 31.5/135.0/3.5 & 20.0/142.5/7.5 \\
        20 (6.6B) & 114.0/48.5/7.5 & 56.0/84.5/29.5 & 42.5/108.0/19.5 & 32.0/117.5/20.5 \\
        24 (7.9B) & 123.0/37.0/10.0 & 70.5/61.5/38.0 & 53.5/80.0/36.5 & 45.5/89.5/35.0 \\
        28 (9.2B) & 131.0/32.0/7.0 & 75.0/63.0/32.0 & 56.0/70.5/43.5 & 46.5/82.5/41.0 \\
        32 (10.4B) & 143.5/21.0/5.5 & 98.0/43.5/28.5 & 73.0/54.0/43.0 & 54.0/65.5/50.5 \\
        36 (11.7B) & 140.5/22.0/7.5 & 98.5/40.5/31.0 & 76.0/49.0/45.0 & 53.0/62.5/54.5 \\
        40 (13B) & 137.5/24.0/8.5 & 102.0/37.0/31.0 & 78.5/45.5/46.0 & 55.0/62.0/53.0 \\ \midrule
        \multicolumn{5}{c}{2 SFT Epochs/4 SoFT Epochs} \\ \midrule
        12 (4.1B)  & 94.5/71.0/4.5 & 44.0/121.0/5.0 & 37.0/130.5/2.5 & 26.5/138.5/5.0 \\
        16 (5.4B)  & 105.0/60.0/5.0 & 55.0/102.0/13.0 & 51.0/110.5/8.5 & 34.0/123.0/13.0 \\
        20 (6.6B)  & 129.5/33.5/7.0 & 73.0/67.5/29.5 & 58.5/85.0/26.5 & 47.0/96.5/26.5 \\
        24 (7.9B)  & 132.0/30.5/7.5 & 89.5/51.0/29.5 & 70.0/62.5/37.5 & 51.0/80.0/39.0 \\
        28 (9.2B)  & 140.0/23.5/6.5 & 89.5/51.0/29.5 & 66.5/60.0/43.5 & 48.5/77.5/44.0 \\
        32 (10.4B)  & 144.5/18.5/7.0 & 103.5/35.0/31.5 & 77.5/52.0/40.5 & 55.5/62.0/52.5 \\
        36 (11.7B)  & 146.0/17.5/6.5 & 105.5/34.5/30.0 & 84.5/44.5/41.0 & 60.0/52.5/57.5 \\
        40 (13B)  & 149.0/15.0/6.0 & 105.0/37.5/27.5 & 87.5/41.5/41.0 & 62.5/53.5/54.0 \\
        \bottomrule
        
    \end{tabular}
    \caption{\small{Pair-wise comparison between Extracted fine-tuned and SoFT sub-models.}}
    \label{tab:pandalm_results_middleLayer_sorted}
\end{table*}

\end{document}